# Learning Partially Observable Deterministic Action Models

**Eyal Amir**                                                                    EYAL@ILLINOIS.EDU
*Computer Science Department*
*University of Illinois, Urbana-Champaign*
*Urbana, IL 61801, USA*

**Allen Chang**                                                          ALLENC256@YAHOO.COM
*2020 Latham st., apartment 25*
*Mountainview, CA 94040, USA*

## Abstract

We present exact algorithms for identifying deterministic-actions' effects and preconditions in dynamic partially observable domains. They apply when one does not know the *action model* (the way actions affect the world) of a domain and must learn it from partial observations over time. Such scenarios are common in real world applications. They are challenging for AI tasks because traditional domain structures that underly tractability (e.g., conditional independence) fail there (e.g., world features become correlated). Our work departs from traditional assumptions about partial observations and action models. In particular, it focuses on problems in which actions are deterministic of simple logical structure and observation models have all features observed with some frequency. We yield tractable algorithms for the modified problem for such domains.

Our algorithms take sequences of partial observations over time as input, and output deterministic action models that could have lead to those observations. The algorithms output all or one of those models (depending on our choice), and are exact in that no model is misclassified given the observations. Our algorithms take polynomial time in the number of time steps and state features for some traditional action classes examined in the AI-planning literature, e.g., *STRIPS* actions. In contrast, traditional approaches for HMMs and Reinforcement Learning are inexact and exponentially intractable for such domains. Our experiments verify the theoretical tractability guarantees, and show that we identify action models exactly. Several applications in planning, autonomous exploration, and adventure-game playing already use these results. They are also promising for probabilistic settings, partially observable reinforcement learning, and diagnosis.

## 1. Introduction

Partially observable domains are common in the real world. They involve situations in which one cannot observe the entire state of the world. Many examples of such situations are available from all walks of life, e.g., the physical worlds (we do not observe the position of items in other rooms), the Internet (we do not observe more than a few web pages at a time), and inter-personal communications (we do not observe the state of mind of our partners).

Autonomous agents' actions involve a special kind of partial observability in such domains. When agents explore a new domain (e.g., one goes into a building or meets a new person), they have limited knowledge about their *action models* (actions' preconditions and effects). These action models do not change with time, but they may depend on state features. Such agents can act intelligently, if they learn how their actions affect the world and use this knowledge to respond to their goals.

Learning action models is important when goals change. When an agent acted for a while, it can use its accumulated knowledge about actions in the domain to make better decisions. Thus, learning action models differs from Reinforcement Learning. It enables reasoning about actions instead of expensive trials in the world.

Learning actions' effects and preconditions is difficult in partially observable domains. The difficulty stems from the absence of useful conditional independence structures in such domains. Most fully observable domains include such structures, e.g., the Markov property (independence of the state at time $t + 1$ from the state at time $t - 1$, given the (observed) state at time $t$). These are fundamental to tractable solutions of learning and decision making.

In partially observable domains those structures fail (e.g., the state of the world at time $t + 1$ depends on the state at time $t - 1$ because we do not observe the state at time $t$), and complex approximate approaches are the only feasible path. For these reasons, much work so far has been limited to fully observable domains (e.g., Wang, 1995; Pasula, Zettlemoyer, & Kaelbling, 2004), hill-climbing (EM) approaches that have unbounded error in deterministic domains (e.g., Ghahramani, 2001; Boyen, Friedman, & Koller, 1999), and approximate action models (Dawsey, Minsker, & Amir, 2007; Hill, Minsker, & Amir, 2007; Kuffner. & LaValle, 2000; Thrun, 2003).

This paper examines the application of an old-new structure to learning in partially observable domains, namely, determinism and logical formulation. It focuses on some such deterministic domains in which tractable learning is feasible, and shows that a traditional assumption about the form of determinism (the STRIPS assumption, generalized to ADL, Pednault, 1989) leads to tractable learning and state estimation. Learning in such domains has immediate applications (e.g., exploration by planning, Shahaf, Chang, & Amir, 2006; Chang & Amir, 2006) and it can also serve as the basis for learning in stochastic domains. Thus, a fundamental advance in the application of such a structure is important for opening the field to new approaches of broader applicability. The following details the technical aspects of our advance.

The main contribution of this paper is an approach called *SLAF* (Simultaneous Learning and Filtering) for exact learning of actions models in partially observable deterministic domains. This approach determines the set of possible transition relations, given an execution sequence of actions and partial observations. For example, the input could come from watching another agent act or from watching the results of our own actions' execution. The approach is *online*, and updates a propositional logical formula called *Transition Belief Formula*. This formula represents the possible transition relations and world states at every time step. In this way, it is similar in spirit to Bayesian learning of HMMs (e.g., Ghahramani, 2001) and Logical Filtering (Amir & Russell, 2003).

The algorithms that we present differ in their range of applicability and their computational complexity. First, we present a deduction-based algorithm that is applicable to any nondeterministic learning problem, but that takes time that is worst-case exponential in the number of domain fluents. Then, we present algorithms that update a logical encoding of all consistent transition relations in polynomial time per step, but that are limited in applicability to special classes of deterministic actions.

One set of polynomial-time algorithms that we present applies to action-learning scenarios in which actions are ADL (Pednault, 1989) (with no conditional effects) and one of the following holds: (a) the action model already has preconditions known, and we observe action failures (e.g., when we perform actions in the domain), or (b) actions execution always succeeds (e.g., when an expert or tutor performs actions).

Our algorithms output a transition belief formula that represents the possible transition relations and states after partial observations of the state and actions. They do so by updating each component of the formula separately in linear time. Thus, updating the transition belief formula with every action execution and observation takes linear time in the size of the input formula.

Processing a sequence of $T$ action executions and observations takes time $O(T^2 \cdot n)$ for case (b). The main reason for this is a linear growth in the representation size of the transition belief formula: at time $t$, the iterative process that updates this formula would process a formula that has size linear in $t$.

For case (a) processing a sequence of length $T$ takes polynomial time $O(T \cdot n^k)$), only if we observe every feature in the domain every $\leq k$ steps in expectation, for some fixed $k$. The reason for this is that the transition belief formula can be kept in $k$-CNF ($k$ *Conjunctive Normal Form*), thus of size $O(n^k)$. (Recall that a propositional formula is in $k$-CNF, if it is of the form $\bigwedge_{i \leq m} \bigvee_{j \leq k} l_{i,j}$, with every $l_{i,j}$ a propositional variable or its negation.) Case (b) takes time $O(T \cdot n)$ under the same assumption.

Another set of polynomial-time algorithms that we present takes linear time in the representation size. In this case actions are known to be injective, i.e., map states 1:1. There, we can bound the computation time for $T$ steps with $O(T \cdot n^k)$, if we approximate the transition-belief formula representation with a $k$-CNF formula.

In contrast, work on learning in Dynamic Bayesian Networks (e.g., Boyen et al., 1999), reinforcement learning in POMDPs (e.g., Littman, 1996), and Inductive Logic Programming (ILP) (e.g., Wang, 1995) either approximate the solution with unbounded error for deterministic domains, or take time $\Omega(2^{2^n})$, and are inapplicable in domains larger than 10 features. Our algorithms are better in this respect, and scale polynomially and practically to domains of 100's of features and more. Section 8 provides a comparison with these and other works.

We conduct a set of experiments that verify these theoretical results. These experiments show that our algorithms are faster and better qualitatively than related approaches. For example, we can learn some ADL actions' effects in domains of $> 100$ features *exactly* and *efficiently*.

An important distinction must be made between learning action models and traditional creation of AI-Planning operators. From the perspective of AI Planning, action models are the result of explicit modeling, taking into account modeling decisions. In contrast, learning action models is deducing all possible transition relations that are compatible with a set of partially observed execution trajectories.

In particular, action preconditions are typically used by the knowledge engineer to control the granularity of the action model so as to leave aside from specification unwanted cases. For example, if driving a truck with insufficient fuel from one site to another might generate unexpected situations that the modeller does not want to consider, then a simple precondition can be used to avoid considering that case. The intention in this paper is not to mimic this modeling perspective, but instead find action models that generate sound states when starting from a sound state. *Sound state* is any state in which the system can be in practice, namely, ones that our observations of real executions can reflect.

Our technical advance for deterministic domains is important for many applications such as automatic software interfaces, internet agents, virtual worlds, and games. Other applications, such as robotics, human-computer interfaces, and program and machine diagnosis can use deterministic action models as approximations. Finally, understanding the deterministic case better can help us develop better results for stochastic domains, e.g., using approaches such as those by Boutilier, Reiter, and Price (2001), Hajishirzi and Amir (2007).

In the following, Section 2 defines SLAF precisely, Section 3 provides a deduction-based exact SLAF algorithm, Section 4 presents tractable action-model-update algorithms, Section 5 gives sufficient conditions and algorithms for keeping the action-model representation compact (thus, overall polynomial time), and Section 7 presents experimental results.

## 2. Simultaneous Learning and Filtering (SLAF)

Simultaneous Learning and Filtering (SLAF) is the problem of tracking a dynamic system from a sequence of time steps and partial observations, when we do not have the system's complete dynamics initially. A solution for SLAF is a representation of all combinations of action models that could possibly have given rise to the observations in the input, and a representation of all the corresponding states in which the system may now be (after the sequence of time steps that were given in the input occurs).

Computing (the solution for) SLAF can be done in a recursive fashion by dynamic programming in which we determine SLAF for time step $t+1$ from our solution of SLAF for time $t$. In this section we define SLAF formally in such a recursive fashion.

Ignoring stochastic information or assumptions, SLAF involves determining the set of possible ways in which actions can change the world (the possible *transition models*, defined formally below) and the set of states the system might be in. Any transition model determines a set of possible states, so a solution to SLAF is a transition model and its associated possible states.

We define SLAF with the following formal tools, borrowing intuitions from work on Bayesian learning of Hidden Markov Models (HMMs) (Ghahramani, 2001) and Logical Filtering (Amir & Russell, 2003).

**Definition 2.1** *A transition system is a tuple* $\langle \mathcal{P}, \mathcal{S}, \mathcal{A}, R \rangle$, *where*

- $\mathcal{P}$ *is a finite set of propositional fluents;*
- $\mathcal{S} \subseteq Pow(\mathcal{P})$ *is the set of world states.*
- $\mathcal{A}$ *is a finite set of actions;*
- $R \subseteq \mathcal{S} \times \mathcal{A} \times \mathcal{S}$ *is the transition relation (*transition model*).*

Thus, a *world state*, $s \in \mathcal{S}$, is a subset of $\mathcal{P}$ that contains propositions true in this state (omitted propositions are false in that state), and $R(s, a, s')$ means that state $s'$ is a possible result of action $a$ in state $s$. Our goal in this paper is to find $R$, given known $\mathcal{P}, \mathcal{S}, \mathcal{A}$, and a sequence of actions and partial observations (logical sentences on any subset of $\mathcal{P}$).

Another, equivalent, representation for $\mathcal{S}$ that we will also use in this paper is the following. A literal is a proposition, $p \in \mathcal{P}$, or its negation, $\neg p$. A *complete term* over $\mathcal{P}$ is a conjunction of literals from $\mathcal{P}$ such that every fluent appears exactly once. Every state corresponds to a complete term of $\mathcal{P}$ and vice versa. For that reason, we sometime identify a state $s \in \mathcal{S}$ with this term. E.g., for states $s_1, s_2, s_1 \vee s_2$ is the disjunction of the complete terms corresponding to $s_1, s_2$, respectively.

A *transition belief state* is a set of tuples $\langle s, R \rangle$ where $s$ is a state and $R$ a transition relation. Let $\mathfrak{R} = Pow(\mathcal{S} \times \mathcal{A} \times \mathcal{S})$ be the set of all possible transition relations on $\mathcal{S}, \mathcal{A}$. Let $\mathfrak{S} = \mathcal{S} \times \mathfrak{R}$. When we hold a transition belief state $\rho \subseteq \mathfrak{S}$ we consider every tuple $\langle s, R \rangle \in \rho$ possible.

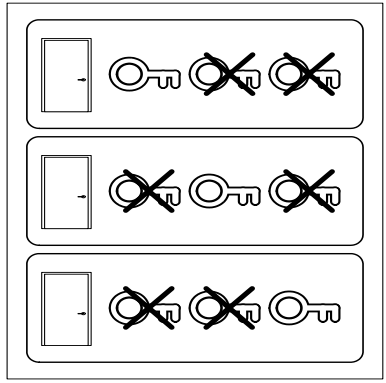

Figure 1: Locked door with unknown key domain.

**Example 2.2** *Consider a domain where an agent is in a room with a locked door (see Figure 1). In its possession are three different keys, and suppose the agent cannot tell from observation only which key opens the door. The goal of the agent is to unlock the door.*

*This domain can be represented as follows: let the set of variables defining the state space be* $\mathcal{P} = \{\textbf{\textit{locked}}\}$ *where* **locked** *is true if and only if the door is locked. Let the set of states be* $\mathcal{S} = \{s_1, s_2\}$ *where* $s_1 = \{\textbf{\textit{locked}}\}$ *(the state in which the door is locked) and* $s_2 = \{\}$ *(here the door is unlocked). Let* $\mathcal{A} = \{\textbf{\textit{unlock}}_1, \textbf{\textit{unlock}}_2, \textbf{\textit{unlock}}_3\}$ *be the three actions wherein the agent tries unlocking the door using each of the three keys.*

*Let* $R_1 = \{\langle s_1, \textbf{\textit{unlock}}_1, s_2\rangle, \langle s_1, \textbf{\textit{unlock}}_2, s_1\rangle, \langle s_1, \textbf{\textit{unlock}}_3, s_1\rangle\}$ *represent a transition relation in which key 1 unlocks the door and the other keys do not. Define* $R_2$ *and* $R_3$ *in a similar fashion (e.g., with* $R_2$ *key 2 unlocks the door and keys 1 and 3 do not). A* transition belief state *represents the set of possibilities that we consider consistent with our observations so far. Consider a transition belief state given by* $\rho = \{\langle s_1, R_1\rangle, \langle s_1, R_2\rangle, \langle s_1, R_3\rangle\}$, *i.e., the state of the world is fully known but the action model is only partially known.*

*We would like the agent to be able to open the door despite not knowing which key opens it. To do this, the agent will learn the actual action model (i.e., which key opens the door). In general, not only will learning an action model be useful in achieving an immediate goal, but such knowledge will be useful as the agent attempts to perform other tasks in the same domain.*□

**Definition 2.3 (SLAF Semantics)** *Let* $\rho \subseteq \mathfrak{S}$ *be a transition belief state. The* SLAF *of* $\rho$ *with actions and observations* $\langle a_j, o_j\rangle_{1 \le j \le t}$ *is defined by*

1. $SLAF[a](\rho) =$
   $\{\langle s', R\rangle \mid \langle s, a, s'\rangle \in R, \ \langle s, R\rangle \in \rho\};$
2. $SLAF[o](\rho) = \{\langle s, R\rangle \in \rho \mid o \text{ is true in } s\};$
3. $SLAF[\langle a_j, o_j\rangle_{i \le j \le t}](\rho) =$
   $SLAF[\langle a_j, o_j\rangle_{i+1 \le j \le t}](SLAF[o_i](SLAF[a_i](\rho))).$

*Step 1 is* progression *with* $a$, *and Step 2* filtering *with* $o$.

**Example 2.4** *Consider the domain from Example 2.2. The progression of* $\rho$ *on the action* **unlock**$_1$ *is given by* $SLAF[\textbf{\textit{unlock}}_1](\rho) = \{\langle s_2, R_1\rangle, \langle s_1, R_2\rangle, \langle s_1, R_3\rangle\}$. *Likewise, the filtering of* $\rho$ *on the observation* ¬**locked** *(the door became unlocked) is given by* $SLAF[\neg \textbf{\textit{locked}}](\rho) = \{\langle s_2, R_1\rangle\}$.□

**Example 2.5** *A slightly more involved example is the following situation presented in Figure 2. There, we have two rooms, a light bulb, a switch, an action of flipping the switch, and an observation, E (we are in the east room). The real states of the world before and after the action, $s2, s2'$, respectively (shown in the top part), are not known to us.*

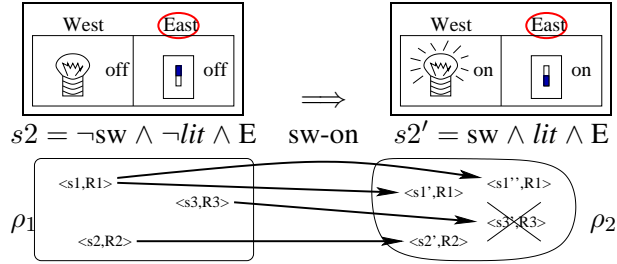

Figure 2: *Top*: Two rooms and flipping the light switch. *Bottom*: SLAF semantics; progressing an action (the arrows map state-transition pairs) and then filtering with an observation (crossing out some pairs).

The bottom of Figure 2 demonstrates how knowledge evolves after performing the action *sw-on*. There, $\rho_1 = \{\langle s_1, R_1 \rangle, \langle s_2, R_2 \rangle, \langle s_3, R_3 \rangle\}$ for some $s_1, R_1, s_3, R_3$, $s_2 = \{E\}$, and $R_2$ that includes $\langle s_2, sw\text{-}on, s_2' \rangle$ (the identity and full details of $R_1, R_2, R_3$ are irrelevant here, so we omit them). $\rho_2$ is the resulting transition belief state after action *sw-on* and observation $E$: $\rho_2 = SLAF[sw\text{-}on, E](\rho_1)$. $\square$

We assume that observations (and an observation model relating observations to state fluents) are given to us as logical sentences over fluents after performing an action. They are denoted with $o$.

This approach to transition belief states generalizes Version Spaces of action models (e.g., Wang, 1995) as follows: If the current state, $s$, is known, then the version space's lattice contains the set of transition relations $\rho^s = \{R \mid \langle s, R \rangle \in \rho\}$. Thus, from the perspective of version spaces, SLAF semantics is equivalent to a set of version spaces, one for each state in which we might be.

This semantics also generalizes belief states: If the transition relation, $R$, is known, then the belief state (set of possible states) is $\rho^R = \{s \mid \langle s, R \rangle \in \rho\}$ (read $\rho$ restricted to $R$), and Logical Filtering (Amir & Russell, 2003) of belief state $\sigma$ and action $a$ is equal to (thus, we can define it as)

$$Filter[a](\sigma) = (SLAF[a](\{\langle s, R \rangle \mid s \in \sigma\}))^R.$$

Thus, SLAF semantics is equivalent to holding a set of belief states, each conditioned on a transition relation, similar to saying "*if the transition relation is $R$, then the belief state (set of states) is $\sigma_R$*".

## 3. Learning by Logical Inference

Learning transition models using Definition 2.3 directly is intractable – it requires space $\Omega(2^{2^{|\mathcal{P}|}})$ in many cases. The reason for that is the explicit representation of the very large set of possible transition-state pairs. Instead, in this section and the rest of this paper we represent transition belief states more compactly using propositional logic. In many scenarios there is some amount of structure that can be exploited to make a propositional representation compact.

A combinatorial argument implies that no encoding is compact for all sets. Nonetheless, we are motivated by the success of propositional (logical) approaches for logical filtering (Amir & Russell, 2003; Shahaf & Amir, 2007) and logical-database regression (Reiter, 1991, 2001), and observe that propositional logic represents compactly some natural sets of exponential size.

In this section we re-define SLAF as an operation on propositional logical formulas with a propositional formula as output. SLAF's input is a propositional formula that represents a transition belief state, and SLAF computes a new transition belief formula from that input and a sequence of actions and observations.

We want to find algorithms for SLAF that manipulate an input formula and produce a correct output. We use general-purpose logical inference for this task in this section. In later sections we sidestep expensive general-purpose inference, and make assumptions that lead to tractable algorithms. For the rest of this paper we focus on *deterministic* transition relations, namely, transition relations that are partial functions (every action has at most one outcome state for every state).

## 3.1 Representing Transition Relations in Logic

Our initial algorithm for solving SLAF (to be presented momentarily) does so with a compact representation of transition belief states. We present this logical encoding of transition belief states first, and define a deduction-based algorithm in the next section.

We use the following general terminology for propositional logical languages (all the terminological conventions apply with or without subscripts and superscripts). $L$ denotes a *vocabulary*, i.e., a set of propositional variables that we use in the present context. $\mathcal{L}$ denotes a *language*, i.e., a set of propositional sentences. $\varphi, \psi$, and other script Greek letters stand for propositional formulas in the language of the present context. $F, G$ also stand for such formulas, but in a restricted context (see below). $L(\varphi)$ denotes the vocabulary of $\varphi$. $\mathcal{L}(L)$ denotes the language built from propositions in $L$ using the standard propositional connectives ($\neg, \vee, \wedge,...$). $\mathcal{L}(\varphi)$ is a shorthand for $\mathcal{L}(L(\varphi))$.

We represent deterministic transition relations with a propositional vocabulary, $L_A$, whose propositions are of the form $a_G^F$, for $a \in \mathcal{A}$, $F$ a literal over $\mathcal{P}$, and $G$ a logical formula. $F$ is the *effect* of $a_G^F$, and $G$ is the *precondition* of $a_G^F$. When proposition $a_G^F$ takes the truth value TRUE, this has the intended meaning that "If $G$ holds in the present state, then $F$ holds in the state that results from executing $a$".

We let $\mathcal{F} \subseteq \mathcal{P} \cup \{\neg p \mid p \in \mathcal{P}\}$ be the set of all effects, $F$, that we consider. We let $\mathcal{G}$ be the set of all preconditions, $G$, that we consider. In the rest of this section and Section 4 we assume that $G$ represents a single state in $\mathcal{S}$. Recall that we identify a state with a complete term which is the conjunction of literals that hold in that state. We use this representation of states and write $a_s^F$ instead of $a_G^F$. Later we build on this definition and consider $G$'s that are more general formulas.

From our assumption ($\mathcal{G} \equiv \mathcal{S}$ for now, as stated above) we conclude that $L_A$ has $O(2^{|\mathcal{P}|} \cdot 2|\mathcal{P}| \cdot |\mathcal{A}|)$ propositional variables. We prove fundamental results for this language and a set of axioms, disregarding the size of the language for a moment. Section 5 focuses on decreasing the language size for computational efficiency.

Our semantics for vocabulary $L_A$ lets every *interpretation* (truth assignment), $M$, for $L_A$ correspond with a transition relation, $R_M$. Every transition relation has at least one (possibly more) interpretation that corresponds to it, so this correspondence is surjective (onto) but not injective (1-to-1). Every propositional sentence $\varphi \in \mathcal{L}(L_A)$ specifies a set of transition models as follows:

The set of *models*[1] (satisfying interpretations) of $\varphi$, $I[\varphi] = \{M \text{ interpretation to } L_A \mid M \models \varphi\}$, specifies the corresponding set of transition relations, $\{R_M \mid M \in I[\varphi]\}$.

Informally, assume that propositions $a_s^{F_1}, ... a_s^{F_k} \in L_A$ take the value TRUE in $M$, and that all other propositions with precondition $s$ take the value FALSE. Then, $R_M$ (with action $a$) takes the state $s$ to a state $s'$ that satisfies $\bigwedge_{i \leq k} F_i$, and is identical to $s$ otherwise. If no such $s'$ exists (e.g., $F_i = \neg F_j$, for some $i, j \leq k$), then $R_M$ takes $s$ to no $s'$ (thus, $a$ is not executable in $s$ according to $R_M$).

The following paragraphs show how interpretations over $L_A$ correspond to transition relations. They culminate with a precise definition of the correspondence between formulas in $\mathcal{L}(L_A \cup \mathcal{P})$ and transition belief states $\rho \subseteq \mathfrak{S}$.

### EVERY INTERPRETATION OF $L_A$ CORRESPONDS TO A UNIQUE TRANSITION RELATION

Every interpretations of $L_A$ defines a unique transition relation $R_M$ as follows. Let $M$ be an interpretation of $L_A$. For every state $s \in \mathcal{S}$ and an action $a \in \mathcal{A}$ we either define a unique state $s'$ such that $\langle s, a, s' \rangle \in R_M$ or decide that there is no $s'$ for which $\langle s, a, s' \rangle \in R_M$.

$M$ gives an interpretation for every proposition $a_s^F$, for $F$ a fluent or its negation. If for any fluent $p \in \mathcal{P}$, $M[a_s^p] = M[a_s^{\neg p}] = TRUE$ ($M[\varphi]$ is the truth value of $\varphi$ according to interpretation $M$), we decide that there is no $s'$ such that $\langle s, a, s' \rangle \in R_M$. Otherwise, define

$$s' = \{p \in \mathcal{P} \mid M \models a_s^p\} \cup \{p \in s \mid M \models \neg a_s^{\neg p}\}$$

In the left-hand side of $\cup$ we consider the cases of $p \in \mathcal{P}$ for which $M[a_s^p] \neq M[a_s^{\neg p}]$, and on the right-hand side of $\cup$ we treat the cases of $p \in \mathcal{P}$ for which $M[a_s^p] = M[a_s^{\neg p}] = FALSE$ (this is called *inertia* because $p$ keeps its previous value for lack of other specifications). Put another way, $s'[p] = M[a_s^p] \vee (s[p] \wedge \neg M[a_s^{\neg p}])$, if we view $s'$ as an interpretation of $\mathcal{P}$. $R_M$ is well defined, i.e., there is only one $R_M$ for every $M$.

### EVERY TRANSITION RELATION HAS AT LEAST ONE CORRESPONDING INTERPRETATION OF $L_A$

It is possible that $R_M = R_{M'}$ for $M \neq M'$. This occurs in two circumstances: (a) cases in which there is no $\langle s, a, s' \rangle \in R_M$ for some $s, a$ and (b) when $M[a_s^p] = M[a_s^{\neg p}] = FALSE$ (inertia) and $M'[a_s^p] = s[p]$, $M'[a_s^{\neg p}] = s[\neg p]$ (not inertia).

For an example of the first circumstance, let $p$ be a fluent, let $M$ be an interpretation such that $M[a_s^p] = M[a_s^{\neg p}]$ for some $G$. Define $M'$ an interpretation that is identical to $M$ on all propositions besides $a_s^p, a_s^{\neg p}$ as follows. Define $M'[a_s^p]$ to have the opposite truth assignment to $M[a_s^p]$ (FALSE instead of TRUE, and TRUE instead of FALSE). Define $M'[a_s^{\neg p}] = M[a_s^{\neg p}]$.

Then, $R_M = R_{M'}$ because they map all pairs $s, a$ in the same way. In particular, for state $s$ that corresponds to $G$, there is no $\langle s, a, s' \rangle \in R_M$ and similarly there is no $\langle s, a, s' \rangle \in R_{M'}$.

Finally, every transition relation $R$ has at least one interpretation $M$ such that $R = R_M$. To see this, define $M_R$ for every $\langle s, a, s' \rangle \in R$ the interpretation to $a_s^p$ ($p$ any fluent) $M_R[a_s^p] = TRUE$ iff $p \in s'$. Also, for the same $\langle s, a, s' \rangle$ define $M_R[a_s^{\neg p}] = FALSE$ iff $p \notin s'$. Finally, for all $s, a$ for which there is no such $s'$, define $M_R[a_s^p] = M_R[a_s^{\neg p}] = TRUE$. Then, $R = R_{M_R}$.

---

1. We overload the word *model* for multiple related meanings. *model* refers to a satisfying interpretation of a logical formula. *Transition model* is defined in Definition 2.1 to be a transition relation in a transition system. *Action model* is define in the Introduction section to be any well-defined specification of actions' preconditions and effects.

### Every Transition Relation Defines a Formula Over $L_A$

Every deterministic transition relation $R$ defines a logical formula whose set of models all map to $R$. There are many such possible formulas, and we define the most general one (up to logical equivalence) that does not make use of inertia.

Define $Th(R)$ as follows.

$$Th_0(R) = \{a_s^F, \neg a_s^{\neg F} \mid a_s^F \in L_A,\ \langle s, a, s' \rangle \in R,\ s' \models F\}$$
$$Th_1(R) = \{a_s^p \vee a_s^{\neg p} \mid p \in \mathcal{P},\ s \in \mathcal{S}\}$$

$$Th_2(R) = \{\bigvee_{p \in \mathcal{P}} (a_s^p \wedge a_s^{\neg p}) \mid \neg \exists s', \langle s, a, s' \rangle \in R\}$$
$$Th(R) = Th_0(R) \cup Th_1(R) \cup Th_2$$

$Th_0$ addresses fluent changes, $Th_1$ addresses fluent inertia (effectively disallowing inertia in our definition), and $Th_2$ addresses conditions in which actions are not executable. Thus, $Th(R)$ includes as a model $M$ every interpretation that satisfies $R_M = R$ and that requires no inertia for its definition of $R_M$. It represents $R$ in that each of its models $M$ satisfies $R_M = R$.

It is illuminating to see how our modeling decisions (above and throughout this section) lead to the last definition. On the one hand, we choose to have *every interpretation* of $L_A$ correspond to a transition relation (we do this to simplify later arguments about logical entailment). Consequently, we associate interpretations $M$ with $M[a_s^F] = M[a_s^{\neg F}] = FALSE$ with transition relations $R(s, a, s')$ that keep the value of $F$ fixed between $s, s'$ (this is *inertia* for $F$ in $a, s$). On the other hand, when we define $Th(R)$ above, we choose axioms that exclude such models (thus, we avoid models that include inertia) because it simplifies our later discussion of learning algorithms.

In summary, we consider every interpretation of $L_A$ as representing exactly one transition relation, and we consider the set of axioms defining $R$ as those that define it directly, i.e., without inertia (without $M[a_s^F] = M[a_s^{\neg F}] = FALSE$).

### Transition Belief States Correspond to Formulas Over $L_A \cup \mathcal{P}$

Thus, for every transition belief state $\rho$ we can define a formula in $\mathcal{L}(L_A \cup \mathcal{P})$ that corresponds to it: $Th(\rho) = \bigvee_{\langle s, R \rangle \in \rho} (s \wedge Th(R))$. Other formulas exist that would characterize $\rho$ in a similar way, and they are not all equivalent. This is so because there are stronger formulas $\varphi \in \mathcal{L}(L_A)$ such that $\varphi \models Th(R)$ and $Th(R) \not\models \varphi$ and every model, $M$, of $\varphi$ satisfies $R_M = R$.

Similarly, for every formula $\varphi \in \mathcal{L}(L_A \cup \mathcal{P})$ we define a transition belief state $\rho(\varphi) = \{\langle M \restriction_\mathcal{P}, R_M \rangle \mid M \models \varphi, \}$, i.e., all the state-transition pairs that satisfy $\varphi$ ($M \restriction_\mathcal{P}$ is $M$ restricted to $\mathcal{P}$, viewed as a complete term over $\mathcal{P}$). We say that formula $\varphi$ is a *transition belief formula*, if $Th(\rho(\varphi)) \equiv \varphi$ (note: $\rho(Th(\rho)) = \rho$ always holds).

### 3.2 Transition-Formula Filtering

In this section, we show that computing the transition belief formula for $SLAF[a](\varphi)$ for successful action $a$ and transition belief formula $\varphi$ is equivalent to a logical consequence finding operation. This characterization of SLAF as consequence-finding permits using consequence finding as an algorithm for SLAF, and is important later in this paper for proving the correctness of our more tractable, specialized algorithms.

Let $Cn^L(\varphi)$ denote the set of logical consequences of $\varphi$ restricted to vocabulary $L$. That is, $Cn^L(\varphi)$ contains the set of prime implicates of $\varphi$ that contain only propositions from the set $L$.

Recall that an *implicate* $\alpha$ of a formula $\varphi$ is a clause entailed by $\varphi$ ($\varphi \models \alpha$). Recall that a *prime implicate* $\alpha$ of a formula $\varphi$ is an implicate of $\varphi$ that is not subsumed (entailed) by any other implicates of $\varphi$.

Consequence finding is any process that computes $Cn^L(\varphi)$ for an input, $\varphi$. For example, propositional *resolution* (Davis & Putnam, 1960; Chang & Lee, 1973) is an efficient consequence finder when used properly (Lee, 1967; del Val, 1999) (Marquis, 2000, surveys results about prime implicates and consequence finding algorithms). Thus, $Cn^L(\varphi) \equiv \{\psi \in \mathcal{L}(L) | \varphi \models \psi\}$.

For a set of propositions $\mathcal{P}$, let $\mathcal{P}'$ represent the same set of propositions but with every proposition primed (i.e., each proposition $f$ is annotated to become $f'$). Typically, we will use a primed fluent to denote the value of the unprimed fluent one step into the future after taking an action. Let $\varphi_{[\mathcal{P}'/\mathcal{P}]}$ denote the same formula as $\varphi$, but with all primed fluents replaced by their unprimed counterparts. For example, the formula $(a \vee b')_{[\mathcal{P}'/\mathcal{P}]}$ is equal to $a \vee b$ when $b \in \mathcal{P}$. (See Section 8 for a discussion and comparison with relevant formal verification techniques.)

The following lemma shows the logical equivalence of existential quantification of quantified boolean formulas and consequence finding restricted to a vocabulary. Recall that *quantified boolean formulas* (QBF) are propositional formulas with the addition of existential and universal quantifiers over propositions. Informally, the QBF $\exists x.\varphi$ is true for a given interpretation if and only if there exists some true/false valuation of $x$ that makes $\varphi$ true under the assignment. The lemma will prove useful for showing the equivalence between SLAF and consequence-finding.

**Lemma 3.1** $\exists x.\varphi \equiv Cn^{L(\varphi) \setminus \{x\}}(\varphi)$, *for any propositional logic formula $\varphi$ and propositional variable $x$.*

PROOF    See Section B.1. $\square$

The lemma extends easily to the case of multiple variables:

**Corollary 3.2** *For any formula $\varphi$ and set of propositional variables $X$, $\exists X.\varphi \equiv Cn^{L(\varphi) \setminus X}(\varphi)$.*

We present an algorithm for updating transition belief formulas whose output is equivalent to that of SLAF (when SLAF is applied to the equivalent transition belief state). Our algorithm applies consequence finding to the input transition belief formula together with a set of axioms that define transitions between time steps. We present this set of axioms first.

For a deterministic (possibly conditional) action, $a$, the action model of $a$ (for time $t$) is axiomatized as

$$\begin{aligned} T_{\mathsf{eff}}(a) = &\bigwedge_{l \in \mathcal{F}, G \in \mathcal{G}} ((a_G^l \wedge G) \Rightarrow l') \;\wedge \\ &\bigwedge_{l \in \mathcal{F}} (l' \Rightarrow (\bigvee_{G \in \mathcal{G}} (a_G^l \wedge G))) \end{aligned} \quad (1)$$

The first part of (1) says that assuming $a$ executes at time $t$, and it causes $l$ when $G$ holds, and $G$ holds at time $t$, then $l$ holds at time $t+1$. The second part says that if $l$ holds after $a$'s execution, then it must be that $a_G^l$ holds and also $G$ holds in the current state. These two parts are very similar to (in fact, somewhat generalize) *effect axioms* and *explanation closure axioms* used in the Situation Calculus (see McCarthy & Hayes, 1969; Reiter, 2001).

Now, we are ready to describe our *zeroth-level algorithm* ($SLAF_0$) for SLAF of a transition belief formula. Let $L' = \mathcal{P}' \cup L_A$ be the vocabulary that includes only fluents of time $t+1$ and effect propositions from $L_A$. Recall (Definition 2.3) that SLAF has two operations: progression (with an action) and filtering (with an observation). At time $t$ we apply progression for the given action $a_t$ and current transition belief formula, $\varphi_t$, and then apply filtering with the current observations:

$$\varphi_{t+1} = SLAF_0[a_t, o_t](\varphi_t) = (Cn^{L'}(\varphi_t \land T_{\mathsf{eff}}(a_t)))_{[\mathcal{P}'/\mathcal{P}]} \land o_t \qquad (2)$$

This is identical to Definition 2.3 (SLAF semantics), with the above replacing 1 and 2.

As stated above, for $SLAF_0$ we can implement $Cn^{L'}(\varphi)$ using consequence finding algorithms such as resolution and some of its variants (e.g., Simon & del Val, 2001; McIlraith & Amir, 2001; Lee, 1967; Iwanuma & Inoue, 2002). The following theorem shows that this formula-SLAF algorithm is correct and exact.

**Theorem 3.3 (Representation)** *For $\varphi$ transition belief formula, $a$ action,*

$$SLAF[a](\{\langle s, R \rangle \in \mathfrak{S} \mid \langle s, R \rangle \text{ satisfies } \varphi\}) =$$
$$\{\langle s, R \rangle \in \mathfrak{S} \mid \langle s, R \rangle \text{ satisfies } SLAF_0[a](\varphi)\}$$

PROOF    See Section B.2. □

This theorem allows us to identify $SLAF_0$ with $SLAF$, and we do so throughout the rest of the paper. In particular, we show that polynomial-time algorithms for SLAF in special cases are correct by showing that their output is logically equivalent to that of $SLAF_0$.

USING THE OUTPUT OF $SLAF_0$

The output of any algorithm for SLAF of a transition belief formula is a logical formula. The way to use this formula for answering questions about SLAF depends on the query and the form of the output formula. When we wish to find if a transition model and state are possible, we wish to see if $M \models \varphi$, for $M$ an interpretation of $L = \mathcal{P} \cup L_A$ and $\varphi$ the output of $SLAF_0$.

The answer can be found by a simple model-checking algorithm[2]. For example, to check that an interpretation satisfies a logical formula we assign the truth values of the variables in the interpretation into the formula; computing the truth value of the formula can be done in linear time.

Thus, this type of query from SLAF takes linear time in the size of the output formula from SLAF because the final query is about a propositional interpretation and a propositional formula.

When we wish to find if a transition model is possible or if a state is possible, we can do so with propositional satisfiability (SAT) solver algorithms (e.g., Moskewicz, Madigan, Zhao, Zhang, & Malik, 2001). Similarly, when we wish to answer whether all the possible models satisfy a property we can use a SAT solver.

**Example 3.4** *Recall Example 2.4 in which we discuss a locked door and three combinations. Let $\varphi_0 = locked$, and let $\varphi_1 = SLAF_0[unlock_2, locked](\varphi_0)$. We wish to find if $\varphi_1$ implies that trying to unlock the door with key 2 fails to open it. This is equivalent to asking if all models consistent with $\varphi_1$ give the value TRUE to $unlock2_{locked}^{locked}$.*

*We can answer such a query by taking the $SLAF_0$ output formula, $\varphi_1$, and checking if $\varphi_1 \land \neg unlock2_{locked}^{locked}$ is SAT (has a model). (Follows from the Deduction Theorem for propositional logic: $\varphi \models \psi$ iff $\varphi \land \neg\psi$ is not SAT.)*

One example application of this approach is the goal achievement algorithm of Chang and Amir (2006). It relies on SAT algorithms to find potential plans given partial knowledge encoded as a transition belief formula.

---

2. This is *not* model checking in the sense used in the Formal Verification literature. There, the model is a transition model, and checking is done by updating a formula in OBDD with some transformations

Our zeroth-level algorithm may enable more compact representation, but it does not guarantee it, nor does it guarantee tractable computation. In fact, no algorithm can maintain compact representation or tractable computation in general. Deciding if a clause is true as a result of SLAF is coNP-hard because the similar decision problem for Logical Filtering is coNP-hard (Eiter & Gottlob, 1992; Amir & Russell, 2003) even for deterministic actions. (The input representation for both problems includes an initial belief state formula in CNF. The input representation for Filtering includes further a propositional encoding in CNF of the (known) transition relation.)

Also, any representation of transition belief states that uses $poly(|\mathcal{P}|)$ propositions grows exponentially (in the number of time steps and $|\mathcal{P}|$) for some starting transition belief states and action sequences, when actions are allowed to be nondeterministic[3]. The question of whether such exponential growth must happen with deterministic actions and *flat formula representations* (e.g., CNF, DNF, etc.; see Darwiche & Marquis, 2002) is open (logical circuits are known to give a solution for deterministic actions, when a representation is given in terms of fluents at time 0, Shahaf et al., 2006).

## 4. Factored Formula Update

Update of any representation is hard when it must consider the set of interactions between all parts of the representation. Operations like those used in $SLAF_0$ consider such interactions, manipulate them, and add many more interactions as a result. When processing can be broken into independent pieces, computation scales up linearly with the number of pieces (i.e., computation time is the total of time it takes for each piece separately). So, it is important to find decompositions that enable such independent pieces and computation. Hereforth we examine one type of decomposition, namely, one that follows logical connectives.

Learning world models is easier when SLAF distributes over logical connectives. A function, $f$, *distributes* over a logical connective, $\cdot \in \{\vee, \wedge, ...\}$, if $f(\varphi \cdot \psi) \equiv f(\varphi) \cdot f(\psi)$. Computation of SLAF becomes tractable, when it distributes over $\wedge, \vee$. The bottleneck of computation in that case becomes computing SLAF for each part separately.

In this section we examine conditions that guarantee such distribution, and present a linear-time algorithm that gives an exact solution in those cases. We will also show that the same algorithm gives a weaker transition belief formula when such distribution is not possible.

Distribution properties that always hold for SLAF follow from set theoretical considerations and Theorem 3.3:

**Theorem 4.1** *For $\varphi, \psi$ transition belief formulas, $a$ action,*

$$SLAF[a](\varphi \vee \psi) \equiv SLAF[a](\varphi) \vee SLAF[a](\psi)$$
$$\models SLAF[a](\varphi \wedge \psi) \Rightarrow SLAF[a](\varphi) \wedge SLAF[a](\psi)$$

PROOF    See Appendix B.3. □

Stronger distribution properties hold for SLAF whenever they hold for Logical Filtering.

**Theorem 4.2** *Let $\rho_1, \rho_2$ be transition belief states.*

$$SLAF[a](\rho_1 \cap \rho_2) = SLAF[a](\rho_1) \cap SLAF[a](\rho_2)$$

---

3. This follows from a theorem about filtering by Amir and Russell (2003), even if we provide a proper axiomatization (note that our axiomatization above is for deterministic actions only).

*iff for every R*

$$Filter[a](\rho_1^R \cap \rho_2^R) = Filter[a](\rho_1^R) \cap Filter[a](\rho_2^R).$$

We conclude the following corollary from Theorems 3.3, 4.2 and theorems by Amir and Russell (2003).

**Corollary 4.3** *For $\varphi, \psi$ transition belief formulas, a action, $SLAF[a](\varphi \wedge \psi) \equiv SLAF[a](\varphi) \wedge SLAF[a](\psi)$ if for every relation R in $\rho_\varphi, \rho_\psi$ one of the following holds:*

1. *a in R maps states 1:1*
2. *a in R has no conditional effects, $\varphi \wedge \psi$ includes all its prime implicates, and we observe if a fails*
3. *The state is known for R: for at most one s, $\langle s, R \rangle \in \rho_\varphi \cup \rho_\psi$.*

Condition 2 combines semantics and syntax. It is particularly useful in the correct computation of SLAF in later sections. It states when $\varphi \wedge \psi$ has a particular syntactic form (namely, together they include their joint prime implicates), and $a$ is simple enough (but not necessarily 1:1), then computation of SLAF can be broken into separate SLAF of $\varphi$ and $\psi$.

Figure 3 presents Procedure Factored-SLAF, which computes SLAF exactly when the conditions of Corollary 4.3 hold. Consequently, Factored-SLAF returns an exact solution whenever our actions are known to be 1:1. If our actions have no conditional effects and their success/failure is observed, then a modified Factored-SLAF can solve this problem exactly too (see Section 5).

---

PROCEDURE Factored-SLAF($\langle a_i, o_i \rangle_{0 < i \le t}, \varphi$)
$\forall i, a_i$ action, $o_i$ observation, $\varphi$ transition belief formula.
    1. For $i$ from 1 to $t$ do,
        (a) Set $\varphi \leftarrow$ Step-SLAF($o_i, a_i, \varphi$).
        (b) Eliminate subsumed clauses in $\varphi$.
    2. Return $\varphi$.

---

PROCEDURE Step-SLAF($o, a, \varphi$)
$o$ an observation formula in $\mathcal{L}(\mathcal{P})$, $a$ an action, $\varphi$ a transition belief formula.
    1. If $\varphi$ is a literal, then return $o \wedge$Literal-SLAF($a, \varphi$).
    2. If $\varphi = \varphi_1 \wedge \varphi_2$, return Step-SLAF($o, a, \varphi_1$)$\wedge$Step-SLAF($o, a, \varphi_2$).
    3. If $\varphi = \varphi_1 \vee \varphi_2$, return Step-SLAF($o, a, \varphi_1$)$\vee$Step-SLAF($o, a, \varphi_2$).

---

PROCEDURE Literal-SLAF($a, \varphi$)
$a$ an action, $\varphi$ a proposition in $L_t$ or its negation.
    1. Return $Cn^{L'}(\varphi \wedge T_{\text{eff}}(a))_{[\mathcal{P}'/P]}$.

---

Figure 3: SLAF using distribution over $\wedge, \vee$

If we pre-compute (and cache) the $2n$ possible responses of Literal-SLAF, then every time step $t$ in this procedure requires linear time in the representation size of $\varphi$, our transition belief formula at that time step. This is a significant improvement over the (super exponential) time taken by a straightforward algorithm, and over the (potentially exponential) time taken by general-purpose consequence finding used in our zeroth-level SLAF procedure above.

**Theorem 4.4** *Step-SLAF$(a, o, \varphi)$ returns a formula $\varphi'$ such that $SLAF[a, o](\varphi) \models \varphi'$. If every run of Literal-SLAF takes time $c$, then Step-SLAF takes time $O(|\varphi|c)$. (recall that $|\varphi|$ is the syntactic, representation size of $\varphi$.) Finally, if we assume one of the assumptions of Corollary 4.3, then $\varphi' \equiv SLAF[a, o](\varphi)$.*

A *belief-state formula* is a transition belief formula that has no effect propositions, i.e., it includes only fluent variables and no propositions of the form $a_G^F$. This is identical to the traditional use of the term *belief-state formula*, e.g., (Amir & Russell, 2003). We can give a closed-form solution for the SLAF of a belief-state formula (procedure Literal-SLAF in Figure 3). This makes procedure Literal-SLAF tractable, avoiding general-purpose inference in filtering a single literal, and also allows us to examine the structure of belief state formulas in more detail.

**Theorem 4.5** *For belief-state formula* $\varphi \in \mathcal{L}(\mathcal{P})$, *action* $a$, $C_a = \bigwedge_{G \in \mathcal{G}, l \in \mathcal{F}} (a_G^l \vee a_G^{-l})$, *and* $G_1, ..., G_m \in \mathcal{G}$ *all the terms in* $\mathcal{G}$ *such that* $G_i \models \varphi$,

$$SLAF[a](\varphi) \equiv \bigwedge_{l_1,...,l_m \in \mathcal{F}} \bigvee_{i=1}^m (l_i \vee a_{G_i}^{-l_i}) \wedge C_a$$

*Here,* $\bigwedge_{l_1,...,l_m \in \mathcal{F}}$ *means a conjunction over all possible (combinations of) selections of* $m$ *literals from* $\mathcal{F}$.

PROOF     See Appendix B.4. □

This theorem is significant in that it says that we can can write down the result of SLAF in a prescribed form. While this form is still potentially of exponential size, it boils down to simple computations. Its proof follows by a straightforward (though a little long) derivation of the possible prime implicates of $SLAF[a](\varphi_t)$.

A consequence of Theorem 4.5 is that we can implement Procedure Literal-SLAF using the equivalence

$$SLAF[a](l) \equiv \left\{ \begin{array}{ll} \text{Theorem 4.5} & L(l) \subseteq \mathcal{P} \\ l \wedge SLAF[a](\textit{TRUE}) & \text{otherwise} \end{array} \right\}$$

Notice that the computation in Theorem 4.5 for $\varphi = l$ a literal is simple because $G_1, ..., G_m$ are all the complete terms in $\mathcal{L}(\mathcal{P})$ that include $l$. This computation does not require a general-purpose consequence finder, and instead we need to answer $2n+1$ queries at the initialization phase, namely, storing in a table the values of $SLAF[a](l)$ for all $l = p$ and $l = \neg p$ for $p \in \mathcal{P}$ and also for $l = \textit{TRUE}$.

In general, $m$ could be as high as $2^{|\mathcal{P}|}$, the number of complete terms in $\mathcal{G}$, and finding $G_1, ..., G_m$ may take time exponential in $|\mathcal{P}|$. Still, the simplicity of this computation and formula provide the basic ingredients needed for efficient computations in the following sections for restricted cases. It should also give guidelines on future developments of SLAF algorithms.

## 5. Compact Model Representation

Previous sections presented algorithms that are potentially intractable for long sequences of actions and observations. For example, in Theorem 4.5, $m$ could be as high as $2^{|\mathcal{P}|}$, the number of complete terms in $\mathcal{G}$. Consequently, clauses may have exponential length (in $n = |\mathcal{P}|$) and there may be a super-exponential number of clauses in this result.

In this section we focus on learning action models efficiently in the presence of action preconditions and failures. This is important for agents that have only partial domain knowledge and are therefore likely to attempt some inexecutable actions.

We restrict our attention to deterministic actions with no conditional effects, and provide an overall polynomial bound on the growth of our representation, its size after many steps, and the time taken to compute the resulting model. This class of actions generalizes STRIPS (Fikes, Hart, & Nilsson, 1972), so our results apply to a large part of the AI-planning literature.

We give an efficient algorithm for learning both non-conditional deterministic action effects and preconditions, as well as an efficient algorithm for learning such actions' effects in the presence of action failures.

## 5.1 Actions of Limited Effect

In many domains we can assume that every action $a$ affects at most $k$ fluents, for some small $k > 0$. It is also common to assume that our actions are STRIPS, and that they may fail without us knowing, leaving the world unchanged. Those assumptions together allow us to progress SLAF with a limited (polynomial factor) growth in the formula size.

We use a language that is similar to the one in Section 3, but which uses only action propositions $a_G^l$ with $G$ being a fluent term of size $k$ (instead of a fluent term of size $n$ in $\mathcal{G}$). Semantically, $a_{l_1 \wedge \ldots \wedge l_k}^l \equiv \bigwedge_{l_{k+1}, \ldots, l_n} a_{l_1 \wedge \ldots \wedge l_n}^l$.

**Theorem 5.1** *Let $\varphi \in \mathcal{L}(\mathcal{P})$ be a belief-state formula, and $a$ a STRIPS action with $\leq k$ fluents affected or in the precondition term. Let $\mathcal{G}^k$ be the set of all terms of $k$ fluents in $\mathcal{L}(\mathcal{P})$ that are consistent with $\varphi$. Then,*

$$SLAF[a](\varphi) \equiv \bigwedge_{\substack{G_1, \ldots, G_k \in \mathcal{G}^k \\ G_1 \wedge \ldots \wedge G_k \models \varphi \\ l_1, \ldots, l_k \in \mathcal{F}}} \bigvee_{i=1}^{k} (l_i \vee a_{G_i}^{\neg l_i}) \wedge C_a$$

*Here, $\bigwedge \ldots$ refers to a conjunction over all possible (combinations of) selections of $m$ literals from $\mathcal{F}$ and $G_1, \ldots, G_k$ from $\mathcal{G}^k$ such that $G_1 \wedge \ldots \wedge G_k \models \varphi$.*

PROOF    See Section B.5. □

The main practical difference between this theorem and Theorem 4.5 is the smaller number of terms that need to be checked for practical computation. The limited language enables and entails a limited number of terms that are at play here. Specifically, when $a$ has at most $k$ literals in its preconditions, we need to check all combinations of $k$ terms $G_1, \ldots, G_k \in \mathcal{G}^k$, computation that is bounded by $O(exp(k))$ iterations.

The proof uses two insights. First, if $a$ has only one case in which change occurs, then every clause in Theorem 4.5 is subsumed by a clause that is entailed by $SLAF[a](\varphi_t)$, has at most one $a_{G_i}^{l_i}$ per literal $l_i$ (i.e., $l_i \neq l_j$ for $i \neq j$) and $G_i$ is a fluent term (has no disjunctions). Second, every $a_G^l$ with $G$ term is equivalent to a formula on $a_{G_i}^l$ with $G_i$ terms of length $k$, if $a$ affects only $k$ fluents.

Thus, we can encode all of the clauses in the conjunction using a subset of the (extended) action effect propositions, $a_G^l$, with $G$ being a term of size $k$. There are $O(n^k)$ such terms, and $O(n^{k+1})$ such propositions. Every clause is of length $2k$, with the identity of the clause determined by the first half (the set of action effect propositions). Consequently, $SLAF[a](\varphi_t)$ takes $O(n^{k^2+k} \cdot k^2)$ space to represent using $O(n^{k^2+k})$ clauses of length $\leq 2k$.

### 5.2 Actions of No Conditional Effects: Revised Language

In this section we reformulate the representation that we presented above in Section 3.1. Let $L_f^0 = \bigcup_{a \in \mathcal{A}} \{a^f, a^{f\circ}, a^{\neg f}, a^{[f]}, a^{[\neg f]}\}$ for every $f \in \mathcal{P}$. Let the vocabulary for the formulas representing transition belief states be defined as $L = \mathcal{P} \cup L_f^0$. The intuition behind propositions in this vocabulary is as follows:

- $a^l$ – "$a$ **causes** $l$" for literal $l$. Formally, $a^l \equiv \bigwedge_{s \in \mathcal{S}} a_s^l$.
- $a^{f\circ}$ – "$a$ **keeps** $f$". Formally, $a^{f\circ} \equiv \bigwedge_{s \in \mathcal{S}} ((s \Rightarrow f) \Rightarrow a_s^f) \wedge ((s \Rightarrow \neg f) \Rightarrow a_s^{\neg f})$.
- $a^{[l]}$ – "$a$ **causes** FALSE **if** $\neg l$". Formally, $a^{[l]} \equiv \bigwedge_{s \in \mathcal{S}} (s \Rightarrow \neg l) \Rightarrow (a_s^l \wedge a_s^{\neg l})$. (Thus, $l$ is a precondition for executing $a$, and it must hold when $a$ executes.)

For each model of a transition belief formula over $L$, the valuation of the fluents from $\mathcal{P}$ defines a state. The valuation of the propositions from all $L_f^0$ defines an unconditional deterministic transition relation as follows: action proposition $a^f$ ($a^{\neg f}$) is true if and only if action $a$ in the transition relation causes $f$ ($\neg f$) to hold after $a$ is executed. Action proposition $a^{f\circ}$ is true if and only if action $a$ does not affect fluent $f$. Action proposition $a^{[f]}$ ($a^{[\neg f]}$) is true if and only if $f$ ($\neg f$) is in the precondition of $a$. We assume the existence of logical axioms that disallow "inconsistent" or "impossible" models. These axioms are:

1. $a^f \vee a^{\neg f} \vee a^{f\circ}$

2. $\neg (a^f \wedge a^{\neg f}) \wedge \neg (a^{\neg f} \wedge a^{f\circ}) \wedge \neg (a^f \wedge a^{f\circ})$

3. $\neg \left( a^{[f]} \wedge a^{[\neg f]} \right)$

for all possible $a \in \mathcal{A}$ and $f \in \mathcal{P}$. The first two axioms state that in every action model, exactly one of $a^f$, $a^{\neg f}$, or $a^{f\circ}$ must hold (thus, $a$ causes $f$, its negation, or keeps $f$ unchanged). The last axiom disallows interpretations where both $a^{[f]}$ and $a^{[\neg f]}$ hold. We state these axioms so that we do not need to represent these constraints explicitly in the transition belief formula itself.

We will use the set theoretic and propositional logic notations for transition belief states interchangeably. Note that the vocabulary we have defined is sufficient for describing any unconditional STRIPS action model, but not any deterministic action model in general.

**Example 5.2** *Consider the domain from Example 2.2. The transition belief state $\varphi$ can be represented by the transition belief formula:*

$$locked \wedge$$
$$((unlock_1^{\neg locked} \wedge unlock_2^{locked\circ} \wedge unlock_3^{locked\circ}) \vee$$
$$(unlock_1^{locked\circ} \wedge unlock_2^{\neg locked} \wedge unlock_3^{locked\circ}) \vee$$
$$(unlock_1^{locked\circ} \wedge unlock_2^{locked\circ} \wedge unlock_3^{\neg locked})).$$

$\square$

We provide an axiomatization that is equivalent to SLAF and is a special case of $T_{\text{eff}}$ (1) with our notation above. We do this over $\mathcal{P}$ and $\mathcal{P}'$. Recall that we intend primed fluents to represent the value of the fluent immediately after action $a$ is taken.

$$\tau_{\mathsf{eff}}(a) \equiv \bigwedge_{f \in \mathcal{P}} \mathrm{Pre}_{a,f} \wedge \mathrm{Eff}_{a,f}$$

$$\mathrm{Pre}_{a,f} \equiv \bigwedge_{l \in \{f, \neg f\}} (a^{[l]} \Rightarrow l)$$

$$\mathrm{Eff}_{a,f} \equiv \bigwedge_{l \in \{f, \neg f\}} ((a^l \vee (a^{f\circ} \wedge l)) \Rightarrow l') \wedge (l' \Rightarrow (a^l \vee (a^{f\circ} \wedge l))).$$

Here $\mathrm{Pre}_{a,f}$ describes the precondition of action $a$. It states that if literal $l$ occurs in the precondition of $a$ then literal $l$ must have held in the state before taking $a$. The formula $\mathrm{Eff}_{a,f}$ describes the effects of action $a$. It states that the fluents before and after taking the action must be consistent according to the action model defined by the propositions $a^f$, $a^{\neg f}$, and $a^{f\circ}$.

Now we can show that the revised axiomatization of action models, $\tau_{\mathsf{eff}}$, leads to an equivalent definition to SLAF within our restricted action models.

**Theorem 5.3** *For any successful action $a$, $SLAF[a](\varphi) \equiv Cn^{L \cup \mathcal{P}'}(\varphi \wedge \tau_{\mathsf{eff}}(a))_{[\mathcal{P}'/\mathcal{P}]}$.*

PROOF     See Appendix B.6. □

## 5.3 Always-Successful Non-Conditional Actions

We are now ready to present an algorithm that learns effects for actions that have no conditional effects. The algorithm allows actions that have preconditions that are not fully known. Still, it assumes that the filtered actions all executed successfully (without failures), so it cannot effectively learn those preconditions (e.g., it would know only some more than it knew originally about those preconditions after seeing a sequence of events). Such a sequence of actions might, for example, be generated by an expert agent whose execution traces can be observed.

The algorithm maintains transition belief formulas in a special *fluent-factored* form, defined below. By maintaining formulas in this special form, we will show that certain logical consequence finding operations can be performed very efficiently. A formula is *fluent-factored*, if it is the conjunction of formulas $\varphi_f$ such that each $\varphi_f$ concerns only one fluent, $f$, and action propositions.

Also, for every fluent, $f$, $\varphi_f$ is conjunction of a positive element, a negative element, and a neutral one $\varphi_f \equiv (\neg f \vee expl_f) \wedge (f \vee expl_{\neg f}) \wedge A_f$, with $expl_f, expl_{\neg f}, A_f$ formulae over action propositions $a^f$, $a^{\neg f}$, $a^{[f]}$, $a^{[\neg f]}$, and $a^{f\circ}$ (possibly multiple different actions). The intuition here is that $expl_f$ and $expl_{\neg f}$ are all the possible explanations for $f$ being true and false, respectively. Also, $A_f$ holds knowledge about actions' effects and preconditions on $f$, knowledge that does not depend on $f$'s current value. Note that any formula in $\mathcal{L}(L_f)$ can be represented as a fluent-factored formula. Nonetheless, a translation sometimes leads to a space blowup, so we maintain our representation in this form by construction.

The new learning algorithm, AS-STRIPS-SLAF[4], is shown in Figure 4. To simplify exposition, it is described for the case of a single action-observation pair, though it should be obvious how to apply the algorithm to sequences of actions and observations. Whenever an action is taken, first the subformulas $A_f$, $expl_f$, and $expl_{\neg f}$ for each $\varphi_f$ are updated according to steps 1.(a)-(c). Then,

---

4. AS-STRIPS-SLAF extends AE-STRIPS-SLAF (Amir, 2005) by allowing preconditions for actions

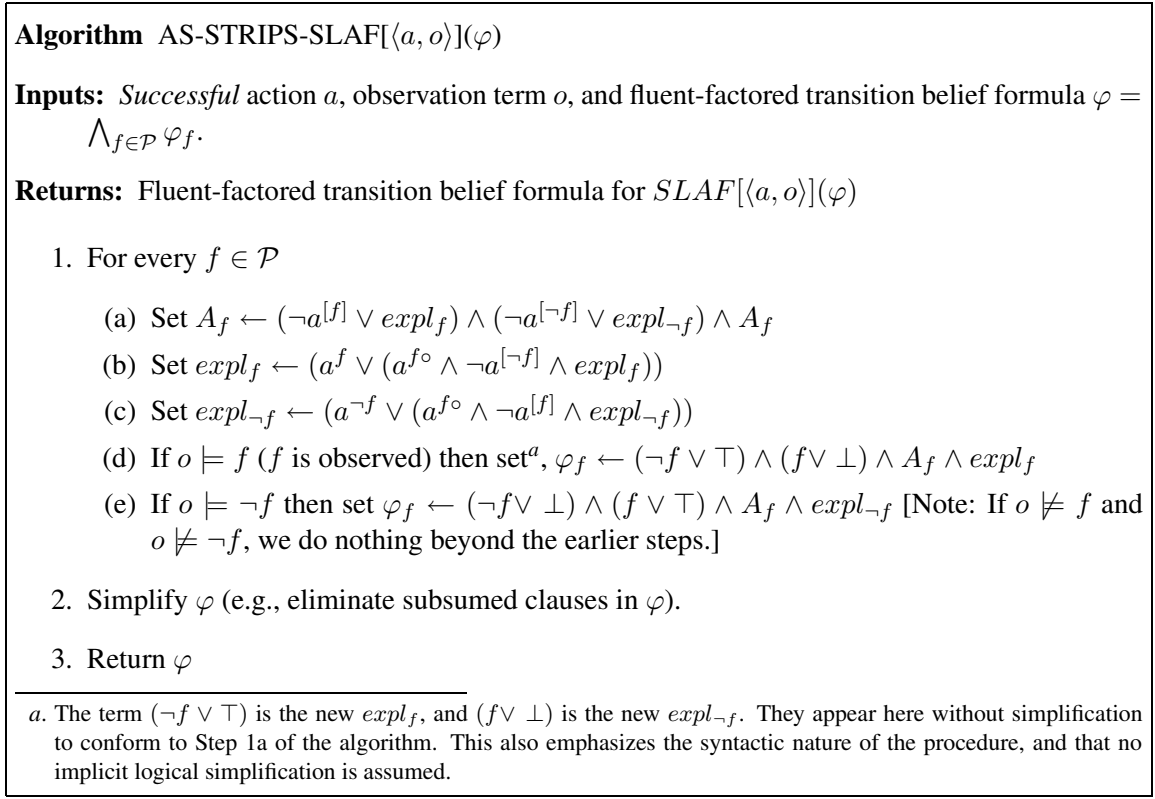

**Algorithm** AS-STRIPS-SLAF$[\langle a, o \rangle](\varphi)$

**Inputs:** *Successful* action $a$, observation term $o$, and fluent-factored transition belief formula $\varphi = \bigwedge_{f \in \mathcal{P}} \varphi_f$.

**Returns:** Fluent-factored transition belief formula for $SLAF[\langle a, o \rangle](\varphi)$

1. For every $f \in \mathcal{P}$

   (a) Set $A_f \leftarrow (\neg a^{[f]} \vee expl_f) \wedge (\neg a^{[\neg f]} \vee expl_{\neg f}) \wedge A_f$

   (b) Set $expl_f \leftarrow (a^f \vee (a^{fo} \wedge \neg a^{[\neg f]} \wedge expl_f))$

   (c) Set $expl_{\neg f} \leftarrow (a^{\neg f} \vee (a^{fo} \wedge \neg a^{[f]} \wedge expl_{\neg f}))$

   (d) If $o \models f$ ($f$ is observed) then set[a], $\varphi_f \leftarrow (\neg f \vee \top) \wedge (f \vee \bot) \wedge A_f \wedge expl_f$

   (e) If $o \models \neg f$ then set $\varphi_f \leftarrow (\neg f \vee \bot) \wedge (f \vee \top) \wedge A_f \wedge expl_{\neg f}$ [Note: If $o \not\models f$ and $o \not\models \neg f$, we do nothing beyond the earlier steps.]

2. Simplify $\varphi$ (e.g., eliminate subsumed clauses in $\varphi$).

3. Return $\varphi$

---

a. The term $(\neg f \vee \top)$ is the new $expl_f$, and $(f \vee \bot)$ is the new $expl_{\neg f}$. They appear here without simplification to conform to Step 1a of the algorithm. This also emphasizes the syntactic nature of the procedure, and that no implicit logical simplification is assumed.

Figure 4: SLAF algorithm for always successful STRIPS actions.

when an observation is received, each $\varphi_f$ is updated according to the observation according to steps 1.(d)-(e). Step 2 merely indicates that in most implementations, it is likely that some simplification procedure will be used on the formula such as subsumption elimination. However, the use of any such simplification procedure is not strictly necessary in order for the theoretical guarantees of our algorithm to hold.

For example, if we know nothing about actions that affect $f$ (e.g., when we start our exploration), then $\varphi_f = (\neg f \vee TRUE) \wedge (f \vee TRUE) \wedge TRUE$. With this representation, $SLAF[a](\varphi_f)$ is the conjunction $(\neg f \vee expl_f) \wedge (f \vee expl_{\neg f}) \wedge A_f$ as computed by step 1 of Procedure AS-STRIPS-SLAF. A similar formula holds for observations.

The following theorem shows the correctness of the algorithm. It shows that the steps taken by the algorithm produce a result equivalent to the logical consequence-finding characterization of SLAF of Theorem 5.3.

**Theorem 5.4** *$SLAF[\langle a, o \rangle](\varphi) \equiv$ AS-STRIPS-SLAF$[\langle a, o \rangle](\varphi)$ for any fluent-factored formula $\varphi$, successfully executed action $a$, and observation term $o$.*

PROOF    See Appendix B.7 □

The time and space complexity of procedure AS-STRIPS-SLAF are given in the following theorem. As a time guarantee, it is shown that the procedure takes linear time in the size of the input formula. Under the condition that the algorithm receives observations often enough—specifically that every fluent is observed at least once every $k$ calls to the procedure—it is possible to show that the transition belief formula remains in $k$-CNF *indefinitely* (recall that $\varphi$ is in $k$-CNF for some fixed $k$, if $\varphi$ is a conjunction of clauses each of size $\leq k$). Thus, regardless of the length of action-observation input sequence, the output of AS-STRIPS-SLAF and the value of $\varphi$ throughout its computation is in $k$-CNF. This amounts to a space guarantee on the size of the formula.

**Theorem 5.5** *The following are true of AS-STRIPS-SLAF:*

1. *The procedure takes linear time in the size of the input formula for a single action, observation pair input.*

2. *If for every fluent and every $k$ steps there is an observation of that fluent in one of those steps, and the input formula is in $k$-CNF, them the resulting formula (after an arbitrary number of steps) is in $k$-CNF.*

3. *If the input of AS-STRIPS-SLAF is fluent-factored, then so is its output.*

PROOF    See Appendix B.8 □

The following corollary follows immediately from the above.

**Corollary 5.6** *In order to process $T$ steps of actions and observations, AS-STRIPS-SLAF requires $O\left(T \cdot |\mathcal{P}|\right)$ time. Additionally, if every fluent is observed every at most $k$ steps, then the resulting formula always has size that is $O\left(|\mathcal{P}| \cdot |\mathcal{A}|^k\right)$.*

This Corollary holds because Theorem 5.5(2) guarantees a bound on the size of our belief-state formula at any point in the algorithm.

## 5.4 Learning Actions Which May Fail

In many partially observable domains, a decision-making agent cannot know beforehand whether each action it decides to take will fail or succeed. In this section we consider possible action failure, and assume that the agent knows whether each action it attempts fails or succeeds after trying the action.

More precisely, we assume that there is an additional fluent $OK$ observed by the agent such that $OK$ is true if and only if the action succeeded. A failed action, in this case, may be viewed as an extra "observation" by the agent that the preconditions for the action were not met. That is, an action failure is equivalent to the observation

$$\neg \bigwedge_{f \in \mathcal{P}} (a^{[f]} \Rightarrow f) \wedge (a^{[\neg f]} \Rightarrow \neg f). \tag{3}$$

Action failures make performing the SLAF operation considerably more difficult. In particular, observations of the form (3) cause interactions between fluents where the value of a particular fluent might no longer depend on only the action propositions for that fluent, but on the action propositions for other fluents as well. Transition belief states can no longer be represented by convenient fluent-factored formulas in such cases, and it becomes more difficult to devise algorithms which give useful time and space performance guarantees.

---

**Algorithm** PRE-STRIPS-SLAF$[a, o](\varphi)$

**Inputs:** Action $a$ and observation term $o$. The transition belief formula $\varphi$ has the following factored form: $\varphi = \bigwedge_i \bigvee_j \varphi_{i,j}$, where each $\varphi_{i,j}$ is a fluent-factored formula.

**Returns:** Filtered transition belief formula $\varphi$

1. If $o \models \neg OK$:

   (a) Set $\varphi \leftarrow \varphi \wedge \bigvee_i F(\neg l_i)$ where $l_i$ are the literals appearing in $a$'s precondition, and $F(l)$ is the fluent-factored formula equivalent to $l$ (i.e., $F(l) = ((l \Rightarrow \top) \wedge (\neg l \Rightarrow \bot) \wedge \top) \wedge \bigwedge_{f \in \mathcal{P}}((f \Rightarrow \top) \wedge (\neg f \Rightarrow \top) \wedge \top))$

   (b) Set $\varphi_{i,j} \leftarrow$ AS-STRIPS-SLAF$[o](\varphi_{i,j})$ for all $\varphi_{i,j}$

2. Else ($o \models OK$):

   (a) For all $\varphi_{i,j}$

      i. Set $\varphi_{i,j} \leftarrow$ AS-STRIPS-SLAF$[P](\varphi_{i,j})$, where $P$ is the precondition of $a$

      ii. Set $\varphi_{i,j} \leftarrow$ AS-STRIPS-SLAF$[\langle a, o \rangle](\varphi_{i,j})$

3. Each $\varphi_{i,j}$ is factored into $A_{i,j} \wedge B_{i,j}$ where $B_{i,j}$ contains all (and only) clauses containing a fluent from $\mathcal{P}$. For any $i$ such that there exists $B$ such that for all $j$, $B_{i,j} \equiv B$, replace $\bigvee_j \varphi_{i,j}$ with $B \wedge \bigvee_j A_{i,j}$

4. Simplify each $\varphi_{i,j}$ (e.g. remove subsumed clauses)

5. Return $\varphi$

---

Figure 5: Algorithm for handling action failures when preconditions are known.

As we shall demonstrate, action failures can be dealt with tractably if we assume that the action preconditions are known by the agent. That is, the agent must learn the effects of the actions it can take, but does not need to learn the preconditions of these actions. In particular, this means that for each action $a$, the algorithm is given access to a formula (more precisely, logical term) $P_a$ describing the precondition of action $a$. Clearly, because the algorithm does not need to learn the preconditions of its actions, we can restrict the action proposition vocabulary used to describe belief states to the ones of the forms $a^f$, $a^{\neg f}$, and $a^{f\circ}$, as we no longer need action propositions of the forms $a^{[f]}$ or $a^{[\neg f]}$.

We present procedure PRE-STRIPS-SLAF[5] (Figure 5) that performs SLAF on transition belief formulas in the presence of action failures for actions of non-conditional effects. It maintains transition belief formulas as conjunctions of disjunctions of fluent-factored formulas (formulas of the form $\varphi = \bigwedge_i \bigvee_j \varphi_{i,j}$ where each $\varphi_{i,j}$ is fluent factored). Naturally, such formulas are a superset of all fluent-factored formulas.

---

5. PRE-STRIPS-SLAF is essentially identical to CNF-SLAF (Shahaf et al., 2006)

The algorithm operates as follows: When an action executes successfully (and an ensuing observation is received), each of the component fluent-factored formulas $\varphi_{i,j}$ is filtered separately according to the AS-STRIPS-SLAF procedure on the action-observation pair (Step 2). On the other hand, when an action fails, a disjunction of fluent-factored formulas is appended to the transition belief formula (Step 1). Each component of the disjunction corresponds to one of the possible reasons the action failed (i.e., to one of the literals occurring in the action's precondition). Finally, as observations are accumulated by the learning algorithm, it collapses disjunctions of fluent-factored formulas occurring in the belief formula together (Step 3) or simplifies them generally (Step 4), decreasing the total size of the formula. As with the case of AS-STRIPS-SLAF, these simplification steps are not necessary in order for our time and space guarantees to hold.

The proof of correctness of Algorithm PRE-STRIPS-SLAF relies on our distribution results from Section 4, Theorem 4.1 and Corollary 4.3.

We proceed to show the correctness of PRE-STRIPS-SLAF. The following theorem shows that the procedure always returns a filtered transition belief formula that is logically weaker than the exact result, so it always produces a safe approximation. Additionally, the theorem shows that under the conditions of Corollary 4.3, the filtered transition belief formula is an exact result.

**Theorem 5.7** *The following are true:*

1. *$SLAF[a, o](\varphi) \models PRE\text{-}STRIPS\text{-}SLAF[a, o](\varphi)$*

2. *$PRE\text{-}STRIPS\text{-}SLAF[a, o](\varphi) \equiv SLAF[a, o](\varphi)$ if Corollary 4.3 holds.*

PROOF    See Appendix B.9. □

Now we consider the time and space complexity of the algorithm. The following theorem shows that (1) the procedure is time efficient, and (2) given frequent enough observations (as in Theorem 5.5), the algorithm is space efficient because the transition belief formula stays indefinitely compact.

**Theorem 5.8** *The following are true of PRE-STRIPS-SLAF:*

1. *The procedure takes time linear in the size of the formula for a single action, observation pair input.*

2. *If every fluent is observed every at most $k$ steps and the input formula is in $m \cdot k$-CNF, then the filtered formula is in $m \cdot k$-CNF, where $m$ is the maximum number of literals in any action precondition.*

PROOF    See Appendix B.10 □

Therefore, we get the following corollary:

**Corollary 5.9** *In order to process $T$ steps of actions and observations, PRE-STRIPS-SLAF requires $O\left(T \cdot |\mathcal{P}|\right)$ time. If every fluent is observed at least as frequently as every $k$ steps, then the resulting formula always has size that is $O\left(|\mathcal{P}| \cdot |\mathcal{A}|^{mk}\right)$.*

## 6. Building on Our Results

In this section we describe briefly how one might extend our approach to include an elaborate observation model, bias, and parametrized actions.

### 6.1 Expressive Observation Model

The observation model that we use throughout this paper is very simple: at every state, if a fluent is observed to have value $v$, then this is its value in the current state. We can consider an observation model that is more general.

An observation model, $O$, is a set of logical sentences that relates propositions in a set $Obs$ with fluents in $\mathcal{P}$. $Obs$ includes propositions that do not appear in $\mathcal{P}$, and which are independent of the previous and following state (times $t-1$ and $t+1$) given the fluents at time $t$.

SLAF with $o$ is the result of conjoining $\varphi_t$ with $Cn^{L_t}(o \wedge O)$, i.e., finding the prime implicates of $o \wedge O$ and conjoining those with $\varphi_t$. We can embed this extension into our SLAF algorithms above, if we can maintain the same structures that those algorithms use. If $O$ is in $k$-CNF and at every step we observe all but (at most) 1 variable, then finding the prime implicates is easy. Embedding this into the transition-belief formula is done by conjunction of these prime implicates with the formula, and removal of subsumed clauses. The resulting formula is still fluent factored, if the input was fluent factored. Then, the algorithms above remain applicable with the same time complexity, by replacing $o_t$ with the prime implicates of $o_t \wedge O_t$.

**Using the Model**   The algorithms we described above provide an exact solution to SLAF, and all the tuples $\langle s, R \rangle$ that are in this solution are consistent with our observations. They compute a solution to SLAF that is represented as a logical formula. We can use a SAT solver (e.g., Moskewicz et al., 2001) to answer queries over this formula, such as checking if it entails $a^f$, for action $a$ and fluent $f$. This would show if in all consistent models action $a$ makes $f$ have the value TRUE.

The number of variables in the result formula is always *independent of $T$*, and is linear in $|\mathcal{P}|$ for some of our algorithms. Therefore, we can use current SAT solvers to treat domains of 1000 features and more.

**Preference and Probabilistic Bias**   Many times we have information that leads us to prefer some possible action models over others. For example, sometimes we can assume that actions change only few fluents, or we suspect that an action (e.g., *open-door*) does not affect some features (e.g., *position*) normally. We can represent such bias using a preference model (e.g., McCarthy, 1986; Ginsberg, 1987) or a probabilistic prior over transition relations (e.g., Robert, Celeux, & Diebolt, 1993).

We can add this bias at the end of our SLAF computation, and get an exact solution if we can compute the effect of this bias together with a logical formula efficiently. Preferential biases were studied before and fit easily with the result of our algorithms (e.g., we can use implementations of Doherty, Lukaszewicz, & Szalas, 1997, for inference with such bias).

Also, algorithms for inference with probabilistic bias and logical sentences are now emerging and can be used here too (Hajishirzi & Amir, 2007). There, the challenge is to not enumerate tentative models explicitly, a challenge that is overcome with some success in the work of Hajishirzi and Amir (2007) for the similar task of filtering. We can use such algorithms to apply probabilistic bias to the resulting logical formula.

For example, given a probabilistic graphical model (e.g., Bayesian Networks) and a set of propositional logical sentences, we can consider the logical sentences as observations. With this approach, a logical sentence $\varphi$ gives rise to a characteristic function $\delta_\varphi(\overrightarrow{x})$ which is 1 when $\overrightarrow{x}$ satisfies $\varphi$ and 0 otherwise. For a conjunction of clauses we get a set of such functions (one per clause). Thus, inference in the combined probabilistic-logical system is a probabilistic inference. For example, one can consider variable elimination (e.g., Dechter, 1999) in which there are additional potential functions.

**Parametrized Actions**   In many systems and situations it is natural to use parametrized actions. These are action schemas whose effect depend on their parameters, but their definition applies identically to all instantiations.

For example, $move(b, x, y)$ can be an action which moves $b$ from position $x$ to position $y$, with $b, x, y$ as parameters of the action. These are very common in planning systems (e.g., STRIPS, PDDL, Situation Calculus). A complete treatment of parameterized actions is outside the scope of this paper, but we give guidelines for a generalization of our current approach for such actions.

Consider a domain with a set of fluent predicates and a universe of named objects. The propositional fluents that are defined in this domain are all ground instantiations of predicate fluents. SLAF can work on this set of propositional fluents and instantiated actions in the same manner as for the rest of this paper. We have action propositions $a(\overrightarrow{X})^l_G$ instantiated for every vector of object names $\overrightarrow{X}$.

The different treatment comes in additional axioms that say that $\forall \overrightarrow{x}, \overrightarrow{y}. a(\overrightarrow{x})^l_G \iff a(\overrightarrow{y})^l_G$. Inference over a transition belief state with these axioms will be able to join information collected about different instantiations of those actions. We expect that a more thorough treatment will be able to provide more efficient algorithms whose time complexity depend on the number of action schemas instead of the number of instantiated actions.

Several approaches already start to address this problem, including the work of Nance, Vogel, and Amir (2006) for filtering and the work of Shahaf and Amir (2006) for SLAF.

## 7. Experimental Evaluation

Previous sections discussed the problem settings that we consider and algorithms for their solutions. They showed that modifying traditional settings for learning in dynamic partially observable domains is important. Determinism alone does not lead to tractability, but our additional assumptions of simple, logical action structure and bounded from 0 frequency of observations for all fluents do. Specifically, so far we showed that the time and space for computing SLAF of a length-$T$ time sequence over $n$ fluents are polynomial in $T$ and $n$.

This section considers the practical considerations involved in using our SLAF procedures. In particular, it examines the following questions:

- How much time and space do SLAF computations take in practice?
- How much time is required to extract a model from the logical formula in the result of our SLAF procedures?
- What is the quality of the learned model (taking an arbitrary consistent model)? How far is it from the true (generating) model?
- Do the conditions for the algorithms' correctness hold in practice?
- Can the learned model be used for successful planning and execution? How do the learning procedures fit with planning and execution?

We implemented our algorithms and ran experiments with AS-STRIPS-SLAF over the following domains taken from the 3rd International Planning Competition (IPC): Drivelog, Zenotravel, Blocksworld, and Depots (details of the domains and the learning results appear in Appendix C).

Each such experiment involves running a chosen algorithm over a sequence of randomly generated action-observation sequences of 5000 steps. Information was recorded every 200 steps.

The random-sequence generator receives the correct description of the domain, specified in PDDL (Ghallab, Howe, Knoblock, McDermott, Ram, Veloso, Weld, & Wilkins, 1998; Fox & Long, 2002) (a plannig-domain description language), the size of the domain, and a starting state. (The size of the domain is the number of propositional fluents in it. It is set by a specification of the number of objects in the domain and the number and arity of predicates in the domain.) It generates a valid sequence of actions and observations for this domain and starting state, i.e., a sequence that is consistent with the input PDDL to that generator but in which actions may fail (action failure is consistent with the PDDL if the action is attempted in a state in which it canot execute).

For our experiments, we chose to have observations as follows: In every time step we select 10 fluents uniformly at random to observe. We applied no additional restrictions (such as making sure each fluent was observed every fixed k steps).

Our SLAF algorithm receives only such a sequences of actions and observations, and no domain information otherwise (e.g., it does not receive the size of the domain, the fluents, the starting state, or the PDDL). The starting knowledge for the algorithm is the empty knowledge, *TRUE*.

For each domain we ran the algorithm over different numbers of propositional fluents (19 to 250 fluents). We collected the time and space taken for each SLAF computation and plotted them as a function of the input-sequence length (dividing by $t$ the total computation time for $t$ steps). The time and space results are shown in Figures 6, 7, 8, and 9. The graphs are broken into the different domains and compare the time and space taken for different domain sizes. The time is SLAF-time without CNF simplification (e.g. we do not remove subsumed clauses)

**How much time and space do SLAF computations take in practice?**   We can answer the first question now. We can observe in these figures that the time per step remains relatively constant throughout execution. Consequently, the time taken to perform SLAF of different domains grows linearly with the number of time steps. Also, we see that the time for SLAF grows with the domain size, but scales easily for moderate domain sizes (1ms per step of SLAF for domains of 200 fluents).

**How much time is required to extract a model from the logical formula in the result of our SLAF procedures?**   Our SLAF procedures return a logical formula for each sequence of actions and observations. We need to apply further work to extract a candidate (consistent) model from that formula. This computation is done with a SAT solver for CNF formulas.

**What is the quality of the learned model (taking an arbitrary consistent model)? How far is it from the true (generating) model?**   There are sometimes many possible models, and with little further bias we must consider each of them possible and as likely. We decided to introduce one such bias, namely, that actions are instances of actions schemas. Thus, the actions are assumed to have the same effect on their parameters (or objects), given properties of those parameters. Thus, actions' effects are assumed independent of the identity of those parameter.

So, with the vanilla implementation, there are propositions which look like

```
((STACK E G) CAUSES (ON E G))
((STACK E G) CAUSES (ON G E))
((STACK A B) CAUSES (ON A B))
((STACK A A) CAUSES (ON A A))
etc.
```

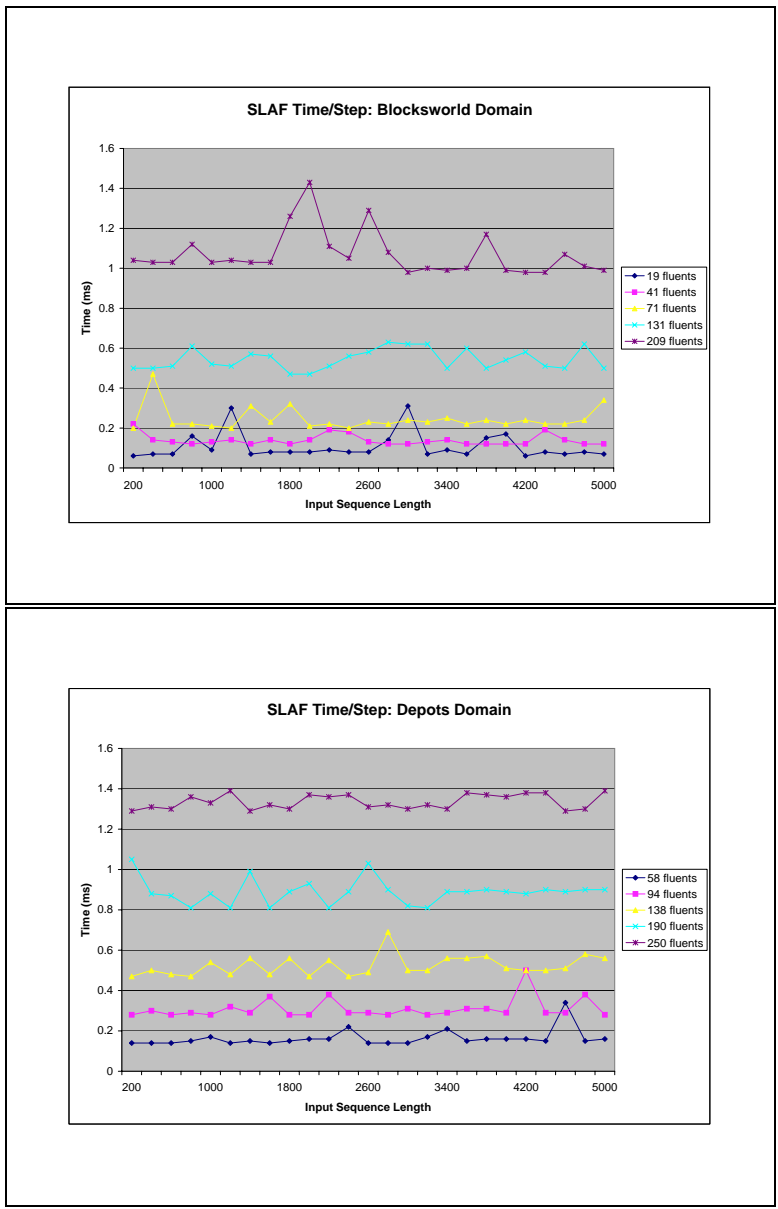

Figure 6: SLAF-time without CNF simplification for domains Blocksworld and Depots

Instead, we replace ground propositions like those above with *schematized* propositions:

```
((STACK ?X ?Y) CAUSES (ON ?X ?Y))
((STACK ?X ?Y) CAUSES (ON ?Y ?X))
((STACK ?X ?Y) CAUSES (ON ?X ?Y))
etc.
```

Thus, the belief-state formula looks something like:

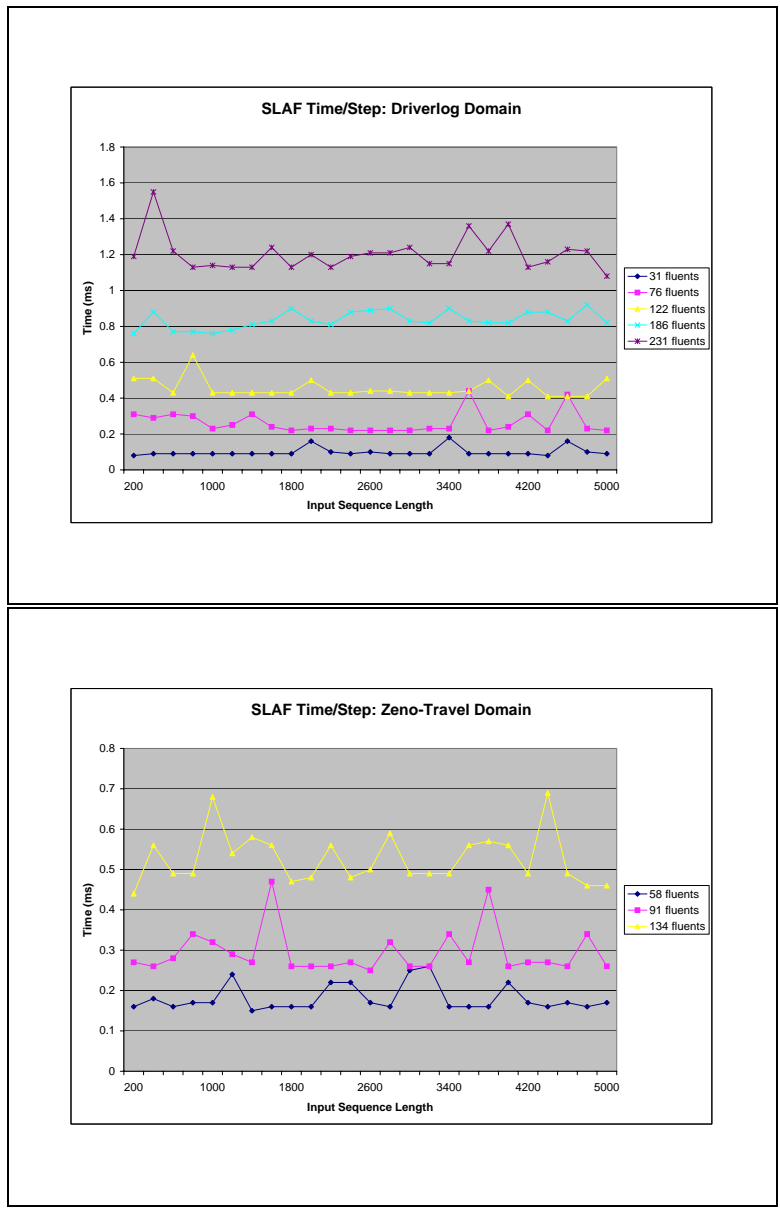

Figure 7: SLAF-time without CNF simplification for domains Driverlog and Zeno-Travel

```
(AND
 (AND
  (OR (ON E G)
   (OR ((STACK ?X ?Y) CAUSES (NOT (ON ?X ?Y)))
    (AND ((STACK ?X ?Y) KEEPS (ON ?X ?Y))
     (NOT ((STACK ?X ?Y) NEEDS (ON ?X ?Y))))))))
  ...
```

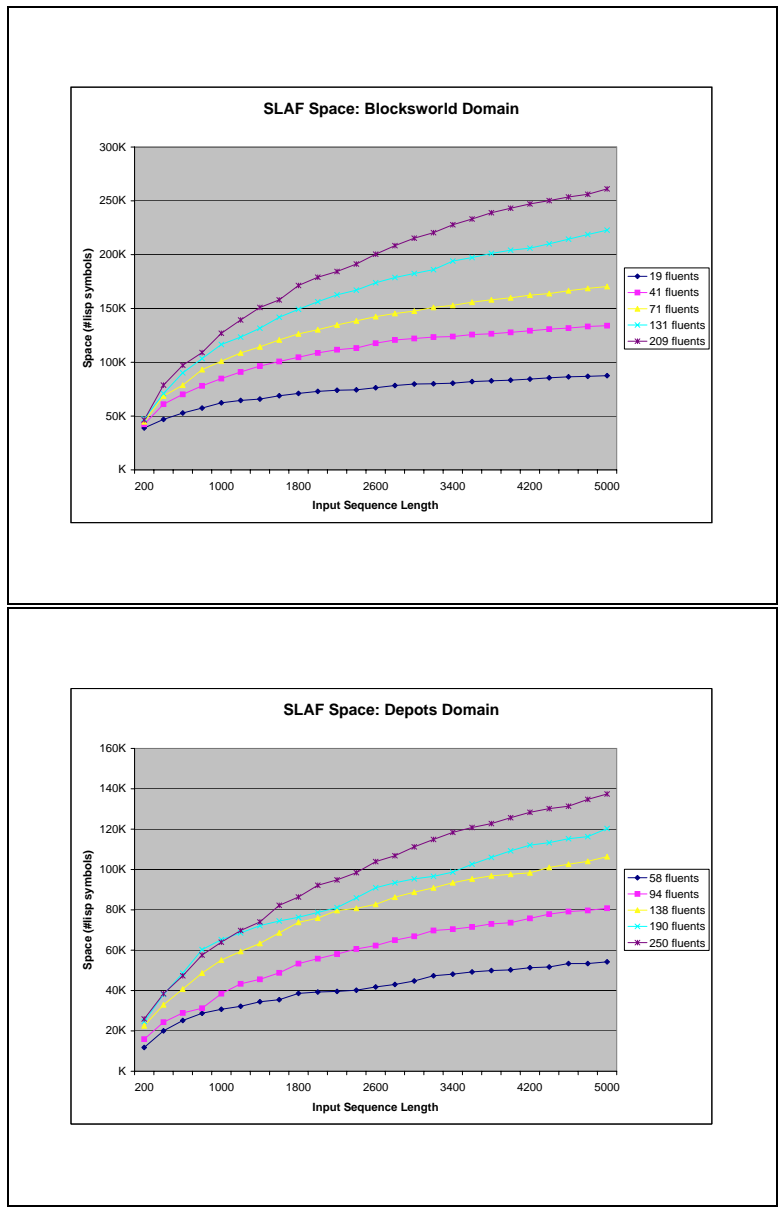

Figure 8: SLAF space for domains Blocksworld and Depots

An example fragment of a model (the complete output is given in Appendix C) that is consistent with our training data is

```
Blocksworld domain:                     converting to CNF
* 209 fluents                           clause count: 235492
* 1000 randomly selected actions        variable count: 187
* 10 fluents observed per step          adding clauses
* "schematized" learning                calling zchaff
```

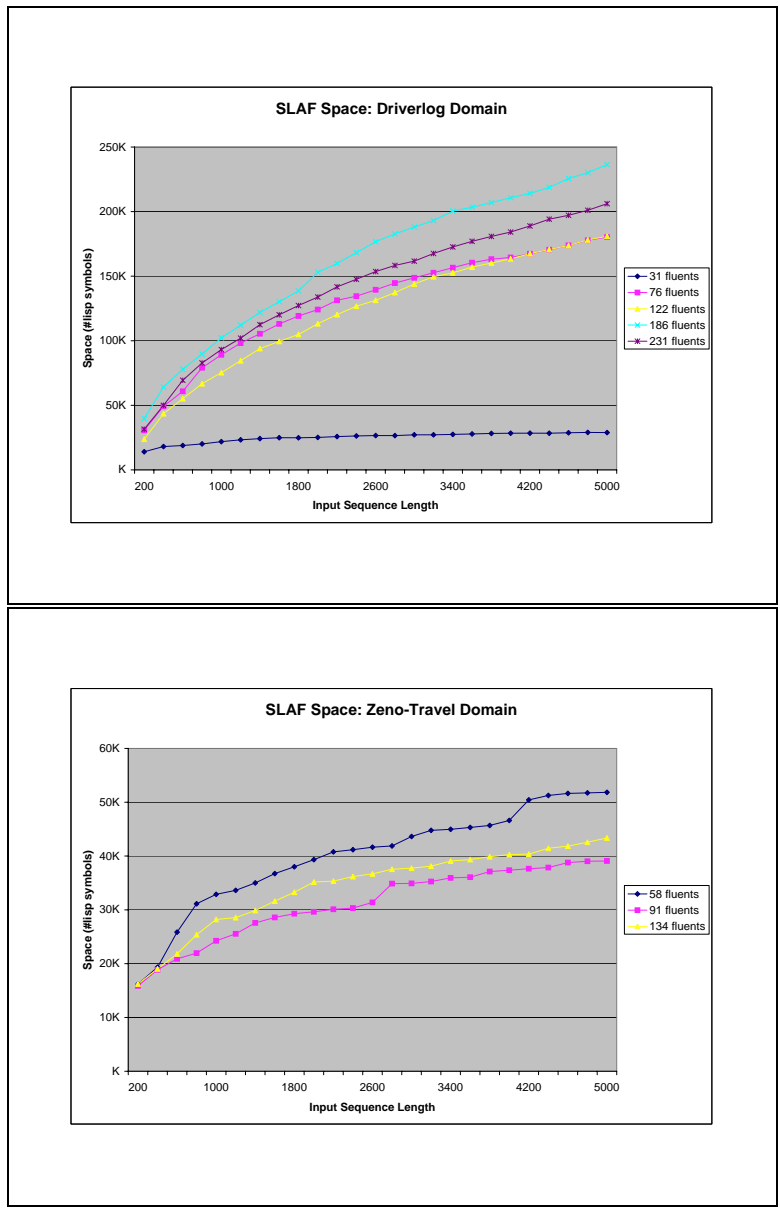

Figure 9: SLAF space for domains Driverlog and Zeno-Travel

```
* 1:1 precondition heuristics            parsing result
                                         SLAF time: 2.203
                                         Inference time: 42.312
                                         Learned model:

(UNSTACK NEEDS (NOT (CLEAR ?UNDEROB)))
(UNSTACK NEEDS (CLEAR ?OB))
(UNSTACK NEEDS (ARM-EMPTY))
```

```
(UNSTACK NEEDS (NOT (HOLDING ?OB)))
(UNSTACK NEEDS (ON ?OB ?UNDEROB))
(UNSTACK CAUSES (CLEAR ?UNDEROB))
(UNSTACK CAUSES (NOT (CLEAR ?OB)))
(UNSTACK CAUSES (NOT (ARM-EMPTY)))
(UNSTACK CAUSES (HOLDING ?OB))
(UNSTACK CAUSES (NOT (ON ?OB ?UNDEROB)))
(UNSTACK KEEPS (ON-TABLE ?UNDEROB))
(UNSTACK KEEPS (ON-TABLE ?OB))
(UNSTACK KEEPS (HOLDING ?UNDEROB))
(UNSTACK KEEPS (ON ?UNDEROB ?UNDEROB))
(UNSTACK KEEPS (ON ?OB ?OB))
(UNSTACK KEEPS (ON ?UNDEROB ?OB))
...
```

Sometimes, there are multiple possible *schematized* propositions that correspond to a ground action proposition, in which case we disjoin the propositions together when doing the replacement (i.e., a single ground propositional symbol gets replaced by a disjunction of schema propositions).

This replacement is a simple procedure, but one that is effective in both deriving more information for fewer steps and also in speeding up model finding from the SLAF formula. We implemented it to run while SLAF runs. One could do this during the SAT-solving portion of the algorithm, with an equivalent result.

Regarding the latter, we ran into scaling issues with the SAT solver (ZChaff , Moskewicz et al., 2001; Tang, Yinlei Yu, & Malik, 2004) and the common lisp compiler in large experiments. We got around these issues by applying the replacement scheme above, thus reducing greatly the number of variables that the SAT solver handles.

Another issue that we ran into is that the SAT solver tended to choose models with blank preconditions (the sequences that we used in these experiments include no action failure, so 'no preconditions' is never eliminated by our algorithm). To add some "bias" to the extracted action model, we added axioms of the following form:

```
(or (not (,a causes ,f)) (,a needs (not ,f)))
(or (not (,a causes (not ,f))) (,a needs ,f))
```

These axioms state that if an action $a$ causes a fluent $f$ to hold, then $a$ requires $f$ to not hold in the precondition (similarly, there is an analogous axiom for $\neg f$). Intuitively, these axioms cause the sat solver to favor 1:1 action models. We got the idea for this heuristic from the work of Wu, Yang, and Jiang (2007), which uses a somewhat similar set of axioms to bias their results in terms of learning preconditions. Clearly, these axioms don't always hold, and in the results, one can see that the learned preconditions are often inaccurate.

Some other inaccuracies in the learned action models are reasonable. For example, if a fluent never changes over the course of an action sequence, the algorithm may infer that an arbitrary action causes that fluent to hold.

**Do the conditions for the algorithms' correctness hold in practice?**  In all of those scenarios that we report here, the conditions that guarantee correctness for our algorithms hold. In our experiments we assumed that the main conditions of the algorithms hold, namely, that actions are deterministic and of few preconditions. We did not enforce having observations for every fluent every (fixed) $k$ steps. The latter condition is not necessery for correctness of the algorithm, but is necessary to guarantee polynomial-time computation. Our experiments verify that this is not necessary in practice, and it indeed be the case that the algorithms can have a polynomial-time guarantee for our modified observation

An earlier work of Hlubocky and Amir (2004) has included a modified version of AS-STRIPS-SLAF in their architecture and tested it on a suite of adventure-game-like virtual environments that are generated at random. These include arbitrary numbers of places, objects of various kinds, and configurations and settings of those. There, an agent's task is to exit a house, starting with no knowledge about the state space, available actions and their effects, or characteristics of objects. Their experiments show that the agent learns the effects of its actions very efficiently. This agent makes decisions using the learned knowledge, and inference with the resulting representation is fast (a fraction of a second per SAT problem in domains including more than 30 object, modes, and locations).

**Can the learned model be used for successful planning and execution? How do the learning procedures fit with planning and execution?** The learned model can be used for planning by translating it to PDDL. However, that model is not always the correct one for the domain, so the plan may not be feasible or may not lead to the required goal. In those cases, we can interleave planning, execution, and learning, as was described in the work of Chang and Amir (2006). There, one finds a short plan with a consistent action model, executes that plan, collects observations, and applies SLAF to those. When plan failure can be detected (e.g., that the goal was not achieved), the results of Chang and Amir (2006) guarantee that the joint planning-execution-learning procedure would reach the goal in a bounded amount of time. That bounded time is in fact linear in the length of the longest plan needed for reaching the goal, and is exponential in the complexity of the action model that we need to learn.

## 8. Comparison with Related Work

HMMs (Boyen & Koller, 1999; Boyen et al., 1999; Murphy, 2002; Ghahramani, 2001) can be used to estimate a stochastic transition model from observations. Initially, we expected to compare our work with the HMM implementation of Murphy (2002), which uses EM (a hill-climbing approach). Unfortunately, HMMs require an explicit representation of the state space, and our smallest domain (31 features) requires a transition matrix of $(2^{31})^2$ entries. This prevents initializing HMMs procedures on any current computer.

*Structure learning* approaches in Dynamic Bayes Nets (DBNs) (e.g., Ghahramani & Jordan, 1997; Friedman, Murphy, & Russell, 1998; Boyen et al., 1999) use EM and additional approximations (e.g., using factoring, variation, or sampling), and are more tractable. However, they are still limited to small domains (e.g., 10 features , Ghahramani & Jordan, 1997; Boyen et al., 1999), and also have unbounded errors in discrete deterministic domains, so are not usable in our settings.

A simple approach for learning transition models was devised in the work of Holmes and Charles Lee Isbell (2006) for deterministic POMDPs. There, the transition and observation models are deterministic. This approach is close to ours in that it represents the hidden state and possible models using a finite structure, a Looping Prediction Suffix Tree. This structure can be seen as related to our representation in that both grow models that relate action histories to possible transition models. In our work here such interactions are realized in the recursive structure of the transition-belief formula built by AS-STRIPS-SLAF, e.g.,

$$(a^f \vee (a^{f\circ} \wedge \neg a^{[\neg f]} \wedge expl_f))$$

where $expl_f$ refers to a similar formula to this that was created in the previous time step.

The main difference that we should draw between our work and that of Holmes and Charles Lee Isbell (2006) is that the latter refers to states explicitly, whereas our work refers to features. Consequently, our representation is (provably) more compact and our procedures scale to larger domains both theoretically and in practice. Furthermore, our procedure provably maintains a reference to all the possible models when data is insufficient to determine a single model, whereas the work of Holmes and Charles Lee Isbell (2006) focuses only on the limit case of enough information for determining a single consistent model. On the down-side, our procedure does not consider stochasticity in the belief state, and this remains an area for further development.

A similar relationship holds between our work and that of Littman, Sutton, and Singh (2002). In that work, a model representation is given with a size linear in the number of states. This model, *predictive state representation* (PSR), is based on action/observation histories and predicts behavior based on those histories. That work prefers a low-dimensional vector basis instead a feature-based representation of states (one of the traditional hallmarks of the Knowledge Representation approach). There is no necessary correspondence between the basis vectors and some intuitive features in the real world necessarily. This enables a representation of the world that is more closely based on behavior.

Learning PSRs (James & Singh, 2004) is nontrivial because one needs to find a good low-dimensional vector basis (the stage called *discovery of tests*). That stage of learning PSRs requires matrices of size $\Omega(n^2)$, for states spaces of size $n$.

Our work advances over that line of work in providing correct results in time that is polylogarithmic in the number of states. Specifically, our work learns (deterministic) transition models in polynomial time in the *state features*, thus taking time $O(poly(\log n))$.

Reinforcement Learning (RL) approaches (Sutton & Barto, 1998; Bertsekas & Tsitsiklis, 1996) compute a mapping between world states and preferred actions. They are highly intractable in partially observable domains (Kaelbling, Littman, & Cassandra, 1998), and approximation (e.g., Kearns, Mansour, & Ng, 2000; Meuleau, Peshkin, Kim, & Kaelbling, 1999; Even-Dar, Kakade, & Mansour, 2005; McCallum, 1995) is practical only for small domains (e.g., 10-20 features) with small horizon time $T$.

In contrast to HMMs, DBNs, and RL, our algorithms are exact and tractable in large domains ($>$ 300 features). We take advantages of properties common to discrete domains, such as determinism, limited effects of actions, and observed factor.

Previous work on learning deterministic action models in the AI-Planning literature assumes *fully observable* deterministic domains. These learn parametrized STRIPS actions using, e.g., version spaces (Gil, 1994; Wang, 1995), general classifiers (Oates & Cohen, 1996), or hill-climbing ILP (Benson, 1995). Recently, the work of Pasula et al. (2004) gave an algorithm that learns stochastic actions with no conditional effects. The work of Schmill, Oates, and Cohen (2000) approximates partial observability by assuming that the world is fully observable. We can apply these to partially observable learning problems (sometimes) by using the space of belief states instead of world states, but this increases the problem size to exponentially, so this is not practical for our problem.

Finally, recent research on learning action models in partially observable domains includes the works of Yang, Wu, and Jiang (2005) and of Shahaf and Amir (2006). In the works of Yang et al.

(2005), example plan traces are encoded as a weighted maximum SAT problem, from which a candidate STRIPS action model is extracted. In general, there may be many possible action models for any given set of example traces, and therefore the approach is by nature *approximate* (in contrast to ours, which always identifies the exact set of possible action models). The work also introduces additional approximations in the form of heuristic rules meant to rule out unlikely action models.

The work of Shahaf and Amir (2006) presents an approach for solving SLAF using logical-circuit encodings of transition belief states. This approach performs tractable SLAF for more general deterministic models than those that we present in Section 5, but it also requires SAT solvers for logical circuits. Such SAT solvers are not optimized nowadays in comparison with CNF SAT solvers, so their overall performance for answering questions from SLAF is lower than ours. Most importantly, the representation given by Shahaf and Amir (2006) grows (linearly) with the number of time steps, and this can still hinder long sequences of actions and observations. In comparison, our transition belief formula has bounded size that is independent of the number of time steps that we track.

Our encoding language, $L_A$ is typical in methods for software and hardware verification and testing. Relevant books and methods (e.g., Clarke, Grumberg, & Peled, 1999) are closely related to this representation, and results that we achieve here are applicable there and vice versa. The main distinction we should draw between our work and that done in Formal Methods (e.g., Model Checking and Bounded Model Checking) is that we are able to conclude size bounds for logical formulas involved in the computation. While OBDDs are used with some success in Model Checking, and CNF representations are used with some success in Bounded Model Checking, they have little bounds for the sizes of formulas in theory or in practice. The conditions available in AI applications are used in the current manuscript to deliver such bounds and yield tractability and scalability results that are both theoretical and of practical significance.

Interestingly, methods that use Linear Temporal Logics (LTL) cannot distinguish between what can happen and what actually happens (Calvanese, Giacomo, & Vardi, 2002). Thus, they cannot consider what causes an occurence. Our method is similar in that it does not consider alternate futures for a state explicitly. However, we use an extended language, namely $L_A$, that makes those alternatives explicit. This allows us to forego the limitations of LTL and produce the needed result.

## 9. Conclusions

We presented a framework for learning the effects and preconditions of deterministic actions in partially observable domains. This approach differs from earlier methods in that it focuses on determining the exact set of consistent action models (earlier methods do not). We showed that in several common situations this can be done exactly in time that is polynomial (sometime linear) in the number of time steps and features. We can add bias and compute an exact solution for large domains (hundreds of features), in many cases. Furthermore, we show that the number of action-observation traces that must be seen before convergence is polynomial in the number of features of the domain. These positive results contrast with the difficulty encountered by many approaches to learning of dynamic models and reinforcement learning in partially observable domains.

The results that we presented are promising for many applications, including reinforcement learning, agents in virtual domains, and HMMs. Already, this work is being applied to autonomous agents in adventure games, and exploration is guided by the transition belief state that we compute and information gain criteria. In the future we plan to extend these results to stochastic domains, and domains with continuous features.

## Acknowledgments

We wish to thank Megan Nance for providing the code and samples for the sequence generator, and also wish to thank Dafna Shahaf for an encouraging collaboration that enhanced our development and understanding of these results. The first author also wishes to acknowledge stimulating discussion with Brian Hlubocky on related topics. We wish to acknowledge support from DAF Air Force Research Laboratory Award FA8750-04-2-0222 (DARPA REAL program). The second author wishes to acknowledge support from a University of Illinois at Urbana-Champaign, College of Engineering fellowship. Finally, the first author acknowledges support from a joint Fellowship (2007-2008) with the Center for Advanced Studies and the Beckman Institute at the University of Illinois Urbana-Champaign.

Earlier versions of the results in this manuscript appeared in conference proceedings (Amir, 2005; Shahaf et al., 2006).

## Appendix A. Our Representation and Domain Descriptions

The transition relation $R_M$ for interpretation $M$ in $L_A$ is defined in Section 3.1 in a way that is similar to the interpretation of Domain Descriptions. Domain Descriptions are a common method for specifying structured deterministic domains (Fagin, Ullman, & Vardi, 1983; Lifschitz, 1990; Pednault, 1989; Gelfond & Lifschitz, 1998). Other methods that are equivalent in our contexts include Successor-State Axioms (Reiter, 2001) and the Fluent Calculus (Thielscher, 1998). Their relevance and influence on our work merit a separate exposition and relationship to our work.

A *domain description* $D$ is a finite set of *effect rules* of the form "$a$ **causes** $F$ **if** $G$" that describe the effects of actions, for $F$ and $G$ being *state formulas* (propositional combinations of fluent names). We say that $F$ is the *head* or *effect* and $G$ is the *precondition* for each such rule. We write "$a$ **causes** $F$", for "$a$ **causes** $F$ **if** TRUE". We denote by $E_D(a)$ the set of effect rules in $D$ for action $a \in \mathcal{A}$. For an effect rule $e$, let $G_e$ be its precondition and $F_e$ its effect. Rule $e$ is *active* in state $s$, if $s \models G_e$ ($s$ taken here as a interpretation in $\mathcal{P}$).

Every domain description $D$ defines a unique transition relation $\mathcal{R}_D(s, a, s')$ as follows.

- Let $F(a, s)$ be the conjunction of effects of rules that are active in $s$ for $a$, i.e., $\bigwedge \{F_e \mid e \in E_D(a), s \models G_e\}$. We set $F(a, s) = TRUE$ when no rule of $a$ is active in $s$.
- Let $I(a, s)$ be the set of fluents that are not affected by $a$ in $s$, i.e., $I(a, s) = \{f \in \mathcal{P}_D \mid \forall e \in E_D(a) (s \models G_e) \Rightarrow (f \notin L(F_e))\}$.
- Define (recalling that world states are sets of fluents)

$$\mathcal{R}_D = \left\{ \langle s, a, s' \rangle \;\middle|\; \begin{array}{l} (s' \cap I(a, s)) = (s \cap I(a, s)) \\ \text{and } s' \models F(a, s) \end{array} \right\} \tag{4}$$

Thus, if action $a$ has an effect of *FALSE* in $s$, then it cannot execute in $s$.

This definition applies inertia (a fluent keeps its value) to all fluents that appear in no active rule. In some contexts it is useful to specify this inertia explicitly by extra effect rules of the form "$a$ **keeps** $f$ if $G$", for a fluent $f \in \mathcal{P}$. It is a shorthand for writing the two rules "$a$ **causes** $f$ if $f \wedge G$" and "$a$ **causes** $\neg f$ if $\neg f \wedge G$". When $D$ includes all its inertia (**keeps**) statements, we say that $D$ is a *complete domain description*.

**Example A.1** *Consider the scenario of Figure 2 and assume that actions and observations occur as in Figure 10. Actions are assumed deterministic, with no conditional effects, and their preconditions must holds before they execute successfully. Then, every action affects every fluent either negatively, positively, or not at all. Thus, every transition relation has a* complete domain description *that includes only rules of the form "a* **causes** *l" or "a* **keeps** *l", where l is a fluent literal (a fluent or its negation).*

| Time step | 1 | | 2 | | 3 | | 4 | | 5 | | 6 |
|---|---|---|---|---|---|---|---|---|---|---|---|
| Action | | go-W | | go-E | | sw-on | | go-W | | go-E | |
| Location | $E$ | | $\neg E$ | | $E$ | | $E$ | | $\neg E$ | | $E$ |
| Bulb | ? | | $\neg lit$ | | ? | | ? | | $lit$ | | ? |
| Switch | $\neg sw$ | | ? | | $\neg sw$ | | $sw$ | | ? | | $sw$ |

Figure 10: An action-observation sequence (table entries are observations). Legend: $E$: east; $\neg E$: west; $lit$: light is on; $\neg lit$: light is off; $sw$: switch is on; $\neg sw$: switch is off.

*Consequently, every transition relation $R$ is completely defined by some domain description $D$ such that (viewing a tuple as a set of its elements)*

$$R \in \prod_{a \in \left\{ \begin{array}{c} go\text{-}W \\ go\text{-}E \\ sw\text{-}on \end{array} \right\}} \left\{ \begin{array}{l} a \textbf{ causes } E, \\ a \textbf{ causes } \neg E \\ a \textbf{ keeps } E \end{array} \right\} \times \left\{ \begin{array}{l} a \textbf{ causes } sw, \\ a \textbf{ causes } \neg sw \\ a \textbf{ keeps } sw \end{array} \right\} \times \left\{ \begin{array}{l} a \textbf{ causes } lit, \\ a \textbf{ causes } \neg lit \\ a \textbf{ keeps } lit \end{array} \right\}$$

*Say that initially we know the effects of go-E, go-W, but do not know what sw-on does. Then, transition filtering starts with the product set of $\mathcal{R}$ (of 27 possible relations) and all possible $2^3$ states. Also, at time step 4 we know that the world state is exactly $\{E, \neg lit, \neg sw\}$. We try sw-on and get that $Filter[sw\text{-}on](\rho_4)$ includes the same set of transition relations but with each of those transition relations projecting the state $\{E, \neg lit, \neg sw\}$ to an appropriate choice from $\mathcal{S}$. When we receive the observations $o_5 = \neg E \wedge \neg sw$ of time step 5, $\rho_5 = Filter[o_5](Filter[sw\text{-}on](\rho_4))$ removes from the transition belief state all the relations that gave rise to $\neg E$ or to $\neg sw$. We are left with transition relations satisfying one of the tuples in*

$$\left\{ \begin{array}{l} sw\text{-}on \textbf{ causes } E, \\ sw\text{-}on \textbf{ keeps } E \end{array} \right\} \times \left\{ sw\text{-}on \textbf{ causes } sw \right\} \times \left\{ \begin{array}{l} sw\text{-}on \textbf{ causes } lit \\ sw\text{-}on \textbf{ causes } \neg lit \\ sw\text{-}on \textbf{ keeps } lit \end{array} \right\}$$

*Finally, when we perform action go-W, again we update the set of states associated with every transition relation in the set of pairs $\rho_5$. When we receive the observations of time step 6, we conclude $\rho_6 = Filter[o_6](Filter[go\text{-}W](\rho_5)) =$*

$$\left\{ \left\langle \left\{ \begin{array}{c} \neg E \\ lit \\ sw \end{array} \right\}, \left\{ \begin{array}{l} sw\text{-}on \textbf{ causes } E, \\ sw\text{-}on \textbf{ causes } sw, \\ sw\text{-}on \textbf{ causes } lit, \\ go\text{-}E... \end{array} \right\} \right\rangle, \left\langle \left\{ \begin{array}{c} \neg E \\ lit \\ sw \end{array} \right\}, \left\{ \begin{array}{l} sw\text{-}on \textbf{ keeps } E, \\ sw\text{-}on \textbf{ causes } sw, \\ sw\text{-}on \textbf{ causes } lit, \\ go\text{-}E... \end{array} \right\} \right\rangle \right\} \quad (5)$$

□

## Appendix B. Proofs

### B.1 Proof of Lemma 3.1: Consequence Finding and Existential Quantification

Proof    Consider some CNF form of $\varphi$. Suppose the clauses containing the literal $x$ are $x \vee \alpha_1, \ldots, x \vee \alpha_a$ where $\alpha_1, \ldots, \alpha_a$ are clauses. Suppose the clauses containing the literal $\neg x$ are $\neg x \vee \beta_1, \ldots, \neg x \vee \beta_b$. Suppose the clauses not containing $x$ or $\neg x$ are $\gamma_1, \ldots, \gamma_c$. Then note that $Cn^{L(\varphi) \backslash \{x\}}(\varphi) \equiv (\bigwedge_{1 \le i \le c} \gamma_i) \wedge (\bigwedge_{1 \le i \le a, 1 \le j \le b} \alpha_i \vee \beta_j)$ (the formula produced by adding all resolvents over variable $x$ and then removing clauses not belonging to $\mathcal{L}(L(\varphi) \backslash \{x\})$), since resolution is complete for consequence finding.

*Necessity.* $(\exists x. \varphi \models Cn^{L(\varphi) \backslash \{x\}}(\varphi))$ Consider any model $m$ of $\exists x. \varphi$. By definition, $m$ can be extended to $L(\varphi)$ (i.e., by assigning some value to $m(x)$) such that $m(\varphi) = 1$. Extend $m$ such that this is the case. Now suppose for contradiction $m$ is not a model of $Cn^{L(\varphi) \backslash \{x\}}(\varphi)$. It cannot be the case that $m(\gamma_k) = 0$ for any $k$, because then $m(\varphi) = 0$, contradiction. Then, it must be the case that $m(\alpha_i \vee \beta_j) = 0$ for some $1 \le i \le a$ and $1 \le j \le b$. Therefore, $m(\alpha_i) = 0$ and $m(\beta_j) = 0$. But $m$ is a model of $\varphi$, so both $m(x \vee \alpha_i) = 1$ or $m(\neg x \vee \beta_j) = 1$. Thus either $m(\alpha_i) = 1$ or $m(\beta_j) = 1$, contradiction.

*Sufficiency.* $(Cn^{L(\varphi) \backslash \{x\}}(\varphi) \models \exists x. \varphi)$ Consider any model $m$ of $Cn^{L(\varphi) \backslash \{x\}}(\varphi)$. Suppose for contradiction $m(\exists x. \varphi) = 0$. That is, if $m$ is extended to $L(\varphi)$, then $m(\varphi) = 0$. Now, extend $m$ to $L(\varphi)$ such that $m(x) = 0$. It cannot be the case that $m(\gamma_k) = 0$ for some $k$ since $m$ models $Cn^{L(\varphi) \backslash \{x\}}(\varphi)$. Because $m(\neg x) = 1$, it cannot be the case that $m(\neg x \vee \beta_j) = 0$ for any $j$. Therefore $m(x \vee \alpha_i) = 0$ for some $1 \le i \le a$. Therefore, $m(\alpha_i) = 0$. Because $m(\alpha_i \vee \beta_j) = 1$ for all $1 \le j \le n$, we must have that $m(\beta_j) = 1$ for all $j$. But now if we alter $m$ such that $m(x) = 1$, then $m$ satisfies $\varphi$, contradiction. $\square$

### B.2 Proof of Theorem 3.3

Proof    Both sides of the equality relation are sets of state-transition-relation pairs. We show that the two sets have the same elements. We show first that the left-hand side of the equality is contained in the right-hand side.

Take $\langle s', R \rangle \in SLAF[a](\{\langle s, R \rangle \in \mathfrak{S} \mid \langle s, R \rangle \text{ satisfies } \varphi\})$. From Definition 2.3 there is $s \in \mathcal{S}$ such that $s \in \{s \in \mathcal{S} \mid \langle s, R \rangle \text{ satisfies } \varphi\}$ such that $\langle s, a, s' \rangle \in R$. In other words, there is $s \in \mathcal{S}$ such that $\langle s, R \rangle$ satisfies $\varphi$ and $\langle s, a, s' \rangle \in R$.

To prove that $\langle s', R \rangle$ satisfies $SLAF_0[a](\varphi)$ we need to show that $\varphi \wedge T_{\text{eff}}(a_t)$ has a model $M$ such that $R_M = R$ and $s'$ interprets $\mathcal{P}'$. Let $M$ be such that $s$ interprets $\mathcal{P}$, $s'$ interprets $\mathcal{P}'$, and $M_R$ interpreting $L_A$ as in the previous section. This interpretation does not satisfy this formula only if one of the conjuncts of $\varphi \wedge \bigwedge_{l \in \mathcal{F}, G \in \mathcal{G}} ((a_G^l \wedge G) \Rightarrow l') \wedge \bigwedge_{l \in \mathcal{F}} (l' \Rightarrow (\bigvee_{G \in \mathcal{G}} (a_G^l \wedge G)))$ is falsified. This cannot be the case for $\varphi$ by our choice of $s$.

Assume by contradiction that $(a_G^l \wedge G) \Rightarrow l'$ fails for some $l$. Then, $(a_G^l \wedge G)$ hold in $M$ and $l'$ is FALSE. The portion of $M$ that interprets $L_A$ is built according to $M_R$, for our $R$. Since $\langle s, R, s' \rangle$ we know that $s'$ satisfies $l$, by the construction of $M_R$. This is a contradicts $l'$ being FALSE in $M$ ($M$ interprets $\mathcal{P}'$ according to $s'$), and therefore we conclude that $(a_G^l \wedge G) \Rightarrow l'$) for every $l'$.

Similarly, assume by contradiction that $(l' \Rightarrow (\bigvee_{G \in \mathcal{G}} (a_G^l \wedge G)))$ fails for some $l$. Then, $l'$ holds in $s'$, and $a_s^l$ fails. Again, from the way we constructed $M_R$ it must be that $a_s^l$ in $M$ takes the value that corresponds to the $l'$'s truth value in $s'$. Thus, it must be that $a_s^l$ takes the value TRUE in $M$, and we are done with the first direction.

For the opposite direction (showing the right-hand side is contained in the left-hand side), take $\langle s', R \rangle \in \mathfrak{S}$ that satisfies $SLAF[a](\varphi)$. We show that

$$\langle s', R \rangle \in SLAF[a](\{\langle s, R \rangle \in \mathfrak{S} \mid \langle s, R \rangle \text{ satisfies } \varphi\}).$$

$\langle s', R \rangle \models SLAF[a](\varphi)$ implies (Corollary 3.2) that there is $s \in \mathcal{S}$ such that $\langle s', R, s \rangle \models \varphi \wedge T_{\text{eff}}(a)$ ($s', R, s$ interpreting $\mathcal{P}', L_A, \mathcal{P}$, respectively). A similar argument to the one we give for the first part shows that $\langle s, a, s' \rangle \in R$, and $\langle s', R \rangle \in SLAF[a](\{\langle s, R \rangle \in \mathfrak{S} \mid \langle s, R \rangle \text{ satisfies } \varphi\})$. $\square$

## B.3 Proof of Theorem 4.1: Distribution of SLAF Over Connectives

For the first part, we will show that the sets of models of $SLAF[a](\varphi \vee \psi)$ and $SLAF[a](\varphi) \vee SLAF[a](\psi)$ are identical.

Take a model $M$ of $SLAF[a](\varphi \vee \psi)$. Let $M'$ be a model of $\varphi \vee \psi$ for which $SLAF[a](M') = M$. Then, $M'$ is a model of $\varphi$ or $M$ is a model of $\psi$. Without loss of generalization, assume $M$ is a model of $\varphi$. Thus, $M \models SLAF[a](\varphi)$, and it follows that $M$ is a model of $SLAF[a](\varphi) \vee SLAF[a](\psi)$.

For the other direction, take $M$ a model of $SLAF[a](\varphi) \vee SLAF[a](\psi)$. Then, $M$ a model of $SLAF[a](\varphi)$ or $M$ is a model of $SLAF[a](\psi)$. Without loss of generalization assume that $M$ is a model of $SLAF[a](\varphi)$. Take $M'$ a model of $\varphi$ such that $M = SLAF[a](M')$. So, $M' \models \varphi \vee \psi$. It follows that $M \models SLAF[a](\varphi \vee \psi)$.

A similar argument follows for the $\wedge$ case. Take a model $M$ of $SLAF[a](\varphi \wedge \psi)$. Let $M'$ be a model of $\varphi \wedge \psi$ for which $SLAF[a](M') = M$. Then, $M'$ is a model of $\varphi$ and $\psi$. Thus, $M \models SLAF[a](\varphi)$ and $M \models SLAF[a](\psi)$. It follows that $M$ is a model of $SLAF[a](\varphi) \wedge SLAF[a](\psi)$. $\square$

## B.4 Proof of Theorem 4.5: Closed form for SLAF of a belief state formula

PROOF SKETCH    We follow the characterization offered by Theorem 3.3 and Formula (1). We take $\varphi_t \wedge a_t \wedge T_{\text{eff}}(a, t)$ and resolve out the literals of time $t$. This resolution is guaranteed to generate a set of consequences that is equivalent to $Cn^{L_{t+1}}(\varphi_t \wedge a_t \wedge T_{\text{eff}}(a, t))$.

Assuming $\varphi_t \wedge a_t$, $T_{\text{eff}}(a, t)$ is logically equivalent to $T_{\text{eff}}(a, t)|_{\varphi} = \bigwedge_{l \in \mathcal{F}, G \in \mathcal{G}, G \models \varphi}((a_G^l \wedge G_t) \Rightarrow l_{t+1}) \wedge \bigwedge_{l \in \mathcal{F}, G \in \mathcal{G}, G \models \varphi}(l_{t+1} \Rightarrow (G_t \Rightarrow a_G^l))$. This follows from two observations. First, notice that $\varphi_t$ implies that for any $G \in \mathcal{G}$ such that $G \not\models \varphi$ we get $G_t \Rightarrow a_G^l$ and $(a_G^l \wedge G_t) \Rightarrow l_{t+1}$ (the antecedent does not hold, so the formula is true). Second, notice that, in the second part of the original $T_{\text{eff}}(a, t)$, $(\bigvee_{G \in \mathcal{G}, G \models \varphi}(a_G^l \wedge G_t))$ is equivalent (assuming $\varphi$) to $(\bigwedge_{G \in \mathcal{G}, G \models \varphi}(G_t \Rightarrow a_G^l))$.

Now, resolving out the literals of time $t$ from $\varphi_t \wedge a_t \wedge T_{\text{eff}}(a, t)|_{\varphi}$ should consider all the resolutions of clauses ($G_t$ is a term) of the form $a_G^l \wedge G_t \Rightarrow l_{t+1}$ and all the clauses of the form $l_{t+1} \Rightarrow (G_t \Rightarrow a_G^l)$ with each other. This yields the equivalent to

$$\bigwedge_{\substack{G_1, \ldots, G_m \in \mathcal{G} \\ \varphi \models \bigvee_{i \leq m} G_i \\ l_1, \ldots, l_m \in \mathcal{F}}} [\bigvee_{i=1}^{m} l_i^{t+1} \vee \bigvee_{i=1}^{m}(a_{G_i}^{\neg l_i} \wedge \neg a_{G_i}^{l_i})]$$

because to eliminate all the literals of time $t$ we have to resolve together sets of clauses with matching $G_i$'s such that $\varphi \models \bigvee_{i \leq n} G_i$. The formula above encodes all the resulting clauses for a chosen set of $G_1, ..., G_m$ and a chosen set of literals $l_1, ..., l_m$. The reason for including $(a_{G_i}^{\neg l_i} \wedge \neg a_{G_i}^{l_i})$ is that we can always choose a clause with $G_i, l_i$ of a specific type (either one that includes $\neg a_G^l$ or one that produces $a_G^{\neg l}$).

Finally, we get the formula in our theorem because $a_G^F \equiv \neg a_G^{\neg F}$ ($G$ characterizes exactly one state in $\mathcal{S}$), and the fact that there is one set of $G_1, ..., G_m$ that is stronger than all the rest (it entails all the rest) because $G_1, ..., G_m$ are complete terms. This set is the complete fluent assignments $G_i$ that satisfy $\varphi$. □

## B.5 Proof of Theorem 5.1: Closed form when $k$ affected fluents

PROOF SKETCH    For literal $l$ and clause $C$ in the conjunction in Theorem 4.5, we aggregate all the action propositions to a single action proposition $a_{G_l}^l$, with $G$ the disjunction of all the complete preconditions for $l$ in $C$ (notice that there cannot be $l$'s negation because then the $C$ is a tautology). Lemma B.2 shows that $G_l$ is equivalent to a fluent term. First we prove the more restricted Lemma B.1.

**Lemma B.1** *Let $C = \bigvee_{i=1}^m (l_i^{t+1} \vee a_{G_i}^{\neg l_i})$ be a clause of the formula in Theorem 4.5, and let $G^l = \bigvee \{G_i \mid l_i = l, \ G_i \models \neg l_i\}$, for any literal $l$[6]. Assume that $a$'s effect is deterministic with only one case with a term precondition (if that case does not hold, then nothing changes). Then, $G^l$ is equivalent to a term.*

PROOF    $G^l$ is a disjunction of complete state terms $G_i$, so it represents the set of states that corresponds to those $G_i$'s. The intuition that we apply is that

$$C \equiv \bigvee_{i=1}^m l_i^{t+1} \vee \neg a_{G_i}^{l_i} \equiv$$
$$(\bigwedge_{i=1}^m a_{G_i}^{l_i}) \Rightarrow (\bigvee_{i=1}^m l_i^{t+1}) \equiv$$
$$a_{G^l}^l \Rightarrow l \vee C'$$

for $C'$ the part of $C$ that does not affect $l$. The reason is that for complete terms $G_i$ we know that $\neg a_{G_i}^{\neg l_i} \equiv a_{G_i}^{l_i}$. Thus, our choice of $G^l$ is such that it includes all the conditions under which $l$ changes, if we assume the precondition $a_{G^l}^l$.

We now compute the action model of $a$ for $l$ by updating a copy of $G^l$. Let $l_i$ be a fluent literal, and set $G_t^l = G^l$.

1. If there is no $G_2^l \in G_t^l$ such that $G_2^l \models l_i$, then all of the terms in $G_t^l$ include $\neg l_i$. Thus, $G_t^l \equiv G_t^l \wedge \neg l_i$, and $a_{G_l}^l \equiv a_{G_t^l \wedge \neg l_i}^l$. Thus, add $\neg l_i$ as a conjunct to $G_t^l$.

2. Otherwise, if there is no $G_2^l \in G_t^l$ such that $G_2^l \models \neg l_i$, then all of the terms in $G_t^l$ include $l_i$. Thus, $G_t^l \equiv G_t^l \wedge l_i$, and $a_{G_l}^l \equiv a_{G_t^l \wedge l_i}^l$. Thus, add $l_i$ as a conjunct to $G_t^l$.

3. Otherwise, we know that under both $l_i$ and $\neg l_i$ the value of the fluent of $l$ changes into $l = TRUE$. Since we assume that **the value of $l$ changes under a single case of its preconditions** (i.e., $a$ has no conditional effects - it either succeeds with the change, or fails with the change), then $l_i$ cannot be part of those preconditions, i.e., for every term $G_i$ in $G_t^l$ we can replace $l_i$ with $\neg l_i$ and vice versa, and the precondition would still entail $l$ after the action. Thus, $a_{G_t^l}^l \equiv a_{G_t^l \setminus \{l_i\}}^l$, for $G_t^l \setminus \{l_i\}$ the result of replacing both $l_i$ and $\neg l_i$ by *TRUE*.

---

6. Not related to the literal $l$ in the proof above.

The result of these replacements is a term $G_t^l$ that is consistent with $\varphi_t$ (it is consistent with the original $G^l$) and satisfies "*a has 1 case*" $\models a_{G_t^l}^l \iff a_{G^l}^l$. $\square$

**Lemma B.2** *Let* $C = \bigvee_{i=1}^m l_i^{t+1} \vee a_{G_i}^{-l_i}$ *be a clause of the formula in Theorem 4.5, and let* $\widehat{G^l} = \bigvee \{G_i \mid l_i = l\}$*, for any literal* $l$*. Assume that* $a$*'s effect is deterministic with only one case with a term precondition (if that case does not hold, then nothing changes). Then, either* $\widehat{G^l}$ *is equivalent to a term, or* $C$ *is subsumed by another clause that is entailed from that formula such that there* $\widehat{G^l}$ *is equivalent to a term.*

PROOF    Consider $G^l$ from Lemma B.1, and let $G_1^l$ a complete fluent term in $\widehat{G^l}$ and not in $G^l$. Thus, $G_1^l \models l$. Let $G_t^l$ the term that is equivalent to $G^l$ according to Lemma B.1.

Clause $C$ is equivalent to $a_{G_t^l}^l \Rightarrow a_{G_1^l}^{-l} \vee \dots$. However, $a_{G_t^l}^l$ already asserts change from $\neg l$ to $l$ in the result of action $a$, and $a_{G_1^l}^{-l}$ asserts a change under a different condition from $l$ to $\neg l$. Thus, "*a has 1 case*" $\models a_{G_t^l}^l \Rightarrow \neg a_{G_1^l}^{-l}$. We get a subsuming clause $C' = C \setminus a_{G_1^l}^{-l}$. In this way we can remove from $C$ all the literals $a_{G_i}^{l_i}$ with $G_i$ not in $G^l$.

After such a process we are left with a clause $C$ that has $G^l \equiv \widehat{G^l}$. However, clause $C$ is now not of the form of Theorem 5.1 because some of the $G_i$'s are missing. How do we represent that in the theorem? We must allow some of the $G_i$'s to be missing.

$\square$

**Proof of theorem continues**    Thus, the representation of $C'$ (the new clause) takes space $O(n^2)$ (each propositional symbol is encoded with $O(n)$ space, assuming constant encoding for every fluent $p \in \mathcal{P}$).

However, the number of propositional symbols is still large (there are $O(2^n)$ fluent terms, so this encoding still requires $O(2^n n)$ (all preconditions to all effects) new propositional symbols. Now we notice that we can limit our attention to clauses $C$ that have at most $k$ literals $l$ whose action proposition $a_{G_l}^l$ satisfies $\models G_l \Rightarrow l$. This is because if we have more than $k$ such propositions in $C$, say $\{a_{G_{l_i}}^{l_i}\}_{i \leq k'}$, then $C \equiv (\bigwedge_{i=1}^k \neg a_{G_{l_i}}^{l_i}) \Rightarrow a_{G_{l_{k+1}}}^{l_{k+1}} \vee \dots \vee_{i \leq m} l_i^{t+1}$, which is always subsumed by $(\bigwedge_{i=1}^k \neg a_{G_{l_i}}^{l_i}) \Rightarrow a_{G_{l_{k+1}}}^{l_{k+1}}$. The latter is a sentence that is always true, if we assume that $a$ can change at most $k$ fluents ($a_{G_{l_{k+1}}}^{l_{k+1}}$ asserts that $l_{k+1}$ remains the same, and $\neg a_{G_{l_i}}^{l_i}$ asserts that $l_i$ changes in one of the conditions satisfied by $G_{l_i}$).

Using clauses of the form $(\bigwedge_{i=1}^k \neg a_{G_{l_i}}^{l_i}) \Rightarrow a_{G_{l_{k+1}}}^{l_{k+1}}$ (which state that we have at most $k$ effects) we can resolve away $a_{G_{l_j}}^{l_j}$ for $j > k$ in every clause $C$ of the original conjunction. Thus, we are left with clauses of the form $C \equiv (\bigwedge_{i=1}^k \neg a_{G_{l_i}}^{l_i}) \Rightarrow \bigvee_{i \leq m} l_i^{t+1}$. Now, since our choice of literals other than $l_1, \dots, l_k$ is independent of the rest of the clause, and we can do so for every polarity of those fluents, we get that all of these clauses resolve (on these combinations of literals) into (and are subsumed by)

$$C \equiv (\bigwedge_{i=1}^k \neg a_{G_{l_i}}^{l_i}) \Rightarrow \bigvee_{i \leq k} l_i^{t+1} \qquad (6)$$

Thus, we get a conjunction of clauses of the form (6), with $G_{l_i}$ ($i \leq k$) being a fluent term. So, the conjunction of clauses in Theorem 4.5 is equivalent to a conjunction of clauses such that each clause has at most $2k$ literals. Thus, the space required to represent each clause is $2kn$.

Finally, we use the fact that every action is dependent on at most $k$ other fluents. Every proposition that asserts no-change in $l_i$ is equivalent to a conjunction of literals stating that none of the possible $k$ preconditions that are consistent with it affect $l_i$. For example $a_{l_1 \wedge \ldots \wedge l_k \wedge l_u}^{l_i}$ implies that $a_{l_1 \wedge \ldots \wedge l_k}^{l_i} \wedge a_{l_1 \wedge \ldots \wedge l_{k-1} \wedge l_u}^{l_i} \wedge \ldots$. Similarly, each one the elements in the conjunction implies $a_{l_1 \wedge \ldots \wedge l_k \wedge l_u}^{l_i}$. $\square$

## B.6 Proof of Theorem 5.3: Equivalent Restricted Definition of Action Axioms

PROOF    Let $\varphi' = \exists \mathcal{P}.(\varphi \wedge \tau_{\mathsf{eff}}(a))$. Now we claim that for any successful action $a$, $SLAF[a](\varphi) \equiv \varphi'_{[\mathcal{P}'/\mathcal{P}]}$. To see this, consider any model of $\varphi'$. The valuation to the fluents in $\bigcup_{f \in \mathcal{F}} L_f^0$ define a transition relation $R$ and the valuation to the fluents in $\mathcal{P}'$ define a state $s' \in \mathcal{S}$ such that $\langle s', R \rangle \in \varphi'$. By the definition of $\varphi'$, $\langle s', R \rangle \in \varphi'$ if and only if there exists $\langle s, R \rangle \in \varphi$ such that $\tau_{\mathsf{eff}}(a)$ is satisfied. Finally note that $\tau_{\mathsf{eff}}(a)$ is satisfied if and only if the preconditions of action $a$ are met and $s'$ is consistent with the effects of action $a$ applied to $s$. That is, $\tau_{\mathsf{eff}}(a)$ is satisfied if and only if $\langle s, a, s' \rangle \in R$. Together, these observations and Corollary 3.2 yield the theorem. $\square$

## B.7 Proof of Theorem 5.4: AS-STRIPS-SLAF is Correct

PROOF    Let the shorthand notation $C(\psi)$ denote $C(\psi) \equiv Cn^{L'}(\psi)_{[\mathcal{P}'/\mathcal{P}]}$.

By definition, $SLAF[\langle a, o \rangle](\varphi) \equiv SLAF[o](SLAF[a](\varphi))$. From Theorem 5.3, we have $SLAF[a](\varphi) \equiv C(\varphi \wedge \tau_{\mathsf{eff}}(a))$. A formula equivalent to $C(\varphi \wedge \tau_{\mathsf{eff}}(a))$ may be generated by resolving out all fluents from $\mathcal{P}$ (by following the procedure from the proof of Lemma 3.1). Suppose $\varphi = \bigwedge_{f \in \mathcal{P}} \varphi_f$ is in fluent-factored form. Then we may rewrite $C(\varphi \wedge \tau_{\mathsf{eff}}(a))$ as:

$$
\begin{aligned}
SLAF[a](\varphi) \quad \equiv \quad & \left( \bigwedge_{f \in \mathcal{P}} C(\mathrm{Pre}_{a,f}) \wedge C(\mathrm{Eff}_{a,f}) \wedge C(\varphi_f) \right) \wedge \qquad (7)\\
& \left( \bigwedge_{f \in \mathcal{P}} C(\mathrm{Pre}_{a,f} \wedge \mathrm{Eff}_{a,f}) \right) \wedge \\
& \left( \bigwedge_{f \in \mathcal{P}} C(\varphi_f \wedge \mathrm{Pre}_{a,f}) \right) \wedge \\
& \left( \bigwedge_{f \in \mathcal{P}} C(\varphi_f \wedge \mathrm{Eff}_{a,f}) \right)
\end{aligned}
$$

This equivalence holds because all resolvents generated by resolving out literals from $\mathcal{P}$ in $C(\varphi \wedge \tau_{\mathsf{eff}}(a))$ will still be generated in the above formula. That is, each pair of clauses that can be possibly resolved together (where a fluent from $\mathcal{P}$ is "resolved out") in $\varphi \wedge \tau_{\mathsf{eff}}(a)$ to generate a new consequence in $C(\varphi \wedge \tau_{\mathsf{eff}}(a))$ will appear together in one of the $C(\cdot)$ components of (7). Because every clause in $\varphi \wedge \tau_{\mathsf{eff}}(a)$ contains at most one literal from $\mathcal{P}$, we see that all possible consequences will be generated.

Now we note that $\mathrm{Eff}_{a,f}$ may be rewritten as follows:

$$
\begin{aligned}
\mathrm{Eff}_{a,f} &\equiv \bigwedge_{l \in \{f, \neg f\}} ((a^l \vee (a^{f \circ} \wedge l)) \Rightarrow l') \wedge (l' \Rightarrow (a^l \vee (a^{f \circ} \wedge l))) \quad (8) \\
&\equiv \bigwedge_{l \in \{f, \neg f\}} (l \Rightarrow (l' \vee a^{\neg l})) \wedge (l' \Rightarrow (a^l \vee a^{f \circ}))
\end{aligned}
$$

It is straightforward to verify the equivalence of the rewritten formula to the original formula. Note that in performing this rewriting, we may discard clauses of the form $a^f \vee a^{\neg f} \vee a^{f \circ}$, as they must be true in every consistent model (given the axioms described in Section 5.2).

Now consider the consequences that are generated by each component of (7). We may compute these consequences by performing resolution. We have that $C(\mathrm{Pre}_{a,f}) \equiv \neg a^{[f]} \vee \neg a^{[\neg f]}$, but we may discard these clauses because only inconsistent action models will violate this clause. By the definition of fluent-factored formulas, $C(\varphi_f) \equiv A_f$. Next, the remaining components can be computed straightforwardly:

$$
\begin{aligned}
C(\mathrm{Eff}_{a,f}) &\equiv \bigwedge_{l \in \{f, \neg f\}} (\neg l' \vee a^l \vee a^{f \circ}) \\
C(\varphi_f \wedge \mathrm{Pre}_{a,f}) &\equiv C(\varphi_f) \wedge C(\mathrm{Pre}_{a,f}) \wedge \bigwedge_{l \in \{f, \neg f\}} (\neg a^{[l]} \vee expl_l) \\
C(\mathrm{Pre}_{a,f} \wedge \mathrm{Eff}_{a,f}) &\equiv C(\mathrm{Pre}_{a,f}) \wedge C(\mathrm{Eff}_{a,f}) \wedge \bigwedge_{l \in \{f, \neg f\}} (\neg l' \vee a^l \vee \neg a^{[\neg l]}) \\
C(\varphi_f \wedge \mathrm{Eff}_{a,f}) &\equiv C(\varphi_f) \wedge C(\mathrm{Eff}_{a,f}) \wedge \bigwedge_{l \in \{f, \neg f\}} (\neg l' \vee a^l \vee expl_l)
\end{aligned}
$$

Finally, it is not difficult to see that in steps (a)-(c) the procedure sets $\varphi$ to the following formula (in fluent-factored form):

$$
SLAF[a](\varphi) \equiv \bigwedge_{f \in \mathcal{P}} A_f \wedge \bigwedge_{l \in \{f, \neg f\}} (\neg a^{[l]} \vee expl_l) \wedge (\neg l' \vee a^l \vee (a^{f \circ} \wedge \neg a^{[\neg l]} \wedge expl_l)
$$

Now, note that $SLAF[o](SLAF[a](\varphi)) \equiv SLAF[a](\varphi) \wedge o$. Note that because $o$ is a term, then $SLAF[a](\varphi) \wedge o$ can be made fluent-factored by performing unit resolution. This is exactly what steps 1.(d)-(e) do. $\square$

## B.8 Proof of Theorem 5.5: AS-STRIPS-SLAF Complexity

PROOF Consider the size of the formula returned by AS-STRIPS-SLAF. *Overview:* We note that if a formula is in $i$-CNF, for any integer $i > 0$, then the filtered formula after *one* step is in $(i + 1)$-CNF. Then, we note that every observation of fluent $f$ *resets* the $f$-part of the belief state formula to 1-CNF (thus, to $i = 1$).

*Further Details:* For the first part, this is because each call to the procedure appends at most one literal to any existing clauses of the formula, and no new clauses of length more than $k + 1$ are generated. Additionally, if every fluent is observed every at most $k$ steps, then the transition belief formula stays in $k$-CNF (i.e., indefinitely compact). This is because existing clauses may only grow in length (1 literal per timestep) when augmented in steps 1.(a)-(c), but when the appropriate fluent is observed in steps 1.(d)-(e), the clauses stop growing. Finally, it is easy to see that each of the steps 1.(a)-(e) can be performed in polynomial time. □

### B.9 Proof of Theorem 5.7: PRE-STRIPS-SLAF is Correct

PROOF    Consider the semantics of SLAF when filtering on a STRIPS action with a known precondition. In the case of an action failure, a world is in the filtered transition belief state if and only if the world did not meet the action precondition (and satisfies the observation). Clearly, step 1 of the algorithm performs this filtering by conjoining the belief formula with the negation of the action precondition (converted into a logically equivalent disjunction of fluent-factored formulas).

In the case of an action success, filtering can be performed by first removing worlds which do not satisfy the action precondition (so in all remaining worlds, the action is executable) and then filtering the remaining worlds using algorithm AS-STRIPS-SLAF. Moreover, by Theorem 4.1 and Corollary 4.3 it follows that filtering the formula $\varphi$ can be performed by filtering each of the subformulas $\varphi_{i,j}$ separately. Furthermore, SLAF$[\langle a, o \rangle](\varphi) \models$ PRE-STRIPS-SLAF$[\langle a, o \rangle](\varphi)$, and PRE-STRIPS-SLAF$[\langle a, o \rangle](\varphi) \equiv$ SLAF$[\langle a, o \rangle](\varphi)$ if any of the conditions of Corollary 4.3. The filtering of each subformula is performed by steps 2 and 3 of the algorithm.

Finally, note that steps 3 and 4 serve only to simplify the belief formula and produce a logically equivalent formula. □

### B.10 Proof of Theorem 5.8: PRE-STRIPS-SLAF Complexity

PROOF    Note that each call to AS-STRIPS-SLAF on a subformula takes time linear in the size of the subformula, and the steps not involving AS-STRIPS-SLAF can be performed in linear time. Thus the the total time complexity is linear. Additionally, note that if every fluent is observed every at most $k$ steps, then every fluent-factored subformula $\varphi_{i,j}$ of the belief formula is in $k$-CNF, by a theorem of Amir (2005). If action preconditions contain at most $m$ literals, then each disjunction of the form $\bigvee_j \varphi_{i,j}$ contains at most $m$ disjuncts. Therefore, the entire belief formula stays in $m \cdot k$-CNF, indefinitely. □

## Appendix C. Experiments And Their Outputs

Our experiments (Section 7) examine properties of our algorithms for learning action models. They show that learning is tractable and exact. In this Appendix section we bring the generating models and the learned models for more detailed comparison by the reader. Recall that our algorithms output a representation for the set of models that are possible given the input. Below we bring only *one* satisfying model from the learned formula.

Our experiments include the following domains from the International Planning Competition (IPC): Drivelog, Zenotravel, Blocksworld, and Depots.

### C.1 Driverlog Domain

The Driverlog domain has the following generating PDDL:

```
(define (domain driverlog)
```

```
    (:requirements :typing)
    (:types        location locatable - object
                   driver truck obj - locatable  )
    (:predicates
          (at ?obj - locatable ?loc - location)
          (in ?obj1 - obj ?obj - truck)
          (driving ?d - driver ?v - truck)
          (path ?x ?y - location)
          (empty ?v - truck) )

(:action LOAD-TRUCK
   :parameters
    (?obj - obj   ?truck - truck   ?loc - location)
   :precondition   (and (at ?truck ?loc) (at ?obj ?loc))
   :effect         (and (not (at ?obj ?loc)) (in ?obj ?truck)))

(:action UNLOAD-TRUCK
   :parameters
    (?obj - obj   ?truck - truck   ?loc - location)
   :precondition   (and (at ?truck ?loc) (in ?obj ?truck))
   :effect         (and (not (in ?obj ?truck)) (at ?obj ?loc)))

(:action BOARD-TRUCK
   :parameters   (?driver - driver   ?truck - truck   ?loc - location)
   :precondition   (and (at ?truck ?loc) (at ?driver ?loc) (empty ?truck))
   :effect   (and (not (at ?driver ?loc)) (driving ?driver ?truck)
                 (not (empty ?truck))))

(:action DISEMBARK-TRUCK
   :parameters   (?driver - driver   ?truck - truck   ?loc - location)
   :precondition   (and (at ?truck ?loc) (driving ?driver ?truck))
   :effect   (and (not (driving ?driver ?truck)) (at ?driver ?loc)
                 (empty ?truck)))

(:action DRIVE-TRUCK
   :parameters   (?truck - truck   ?loc-from - location   ?loc-to -
location
                 ?driver - driver)
   :precondition   (and (at ?truck ?loc-from)   (driving ?driver ?truck)
                 (path ?loc-from ?loc-to))
   :effect   (and (not (at ?truck ?loc-from)) (at ?truck ?loc-to)))

(:action WALK
   :parameters
    (?driver - driver   ?loc-from - location   ?loc-to - location)
   :precondition   (and (at ?driver ?loc-from) (path ?loc-from ?loc-to))
   :effect   (and (not (at ?driver ?loc-from)) (at ?driver ?loc-to)) )
```

One learned model (one possible satisfying model of our formula) from our random-sequence input in this Driverlog domain is the following (brought together with the experimental parameters).

```
Driverlog domain:
* IPC3 problem 99
* 231 fluents
* 1000 randomly selected actions
* 10 fluents observed per step
```

```
* "schematized" learning
* 1:1 precondition heuristics
* Action distribution:
((BOARD-TRUCK . 52) (DRIVE-TRUCK . 86) (DISEMBARK-TRUCK . 52)
 (WALK . 529) (UNLOAD-TRUCK . 139) (LOAD-TRUCK . 142))

converting to CNF
clause count: 82338
variable count: 210
adding clauses
calling zchaff
parsing result
SLAF time: 2.469
Inference time: 8.406
Learned model:

(WALK NEEDS (AT ?DRIVER ?LOC-FROM))
(WALK NEEDS (NOT (AT ?DRIVER ?LOC-TO)))
(WALK CAUSES (AT ?DRIVER ?LOC-TO))
(WALK CAUSES (NOT (AT ?DRIVER ?LOC-FROM)))
(WALK KEEPS (PATH ?LOC-FROM ?LOC-FROM))
(WALK KEEPS (PATH ?LOC-TO ?LOC-TO))
(WALK KEEPS (PATH ?LOC-TO ?LOC-FROM))
(WALK KEEPS (PATH ?LOC-FROM ?LOC-TO))
(DRIVE-TRUCK NEEDS (NOT (AT ?TRUCK ?LOC-TO)))
(DRIVE-TRUCK NEEDS (AT ?TRUCK ?LOC-FROM))
(DRIVE-TRUCK CAUSES (AT ?TRUCK ?LOC-TO))
(DRIVE-TRUCK CAUSES (NOT (AT ?TRUCK ?LOC-FROM)))
(DRIVE-TRUCK KEEPS (AT ?DRIVER ?LOC-TO))
(DRIVE-TRUCK KEEPS (AT ?DRIVER ?LOC-FROM))
(DRIVE-TRUCK KEEPS (DRIVING ?DRIVER ?TRUCK))
(DRIVE-TRUCK KEEPS (PATH ?LOC-TO ?LOC-TO))
(DRIVE-TRUCK KEEPS (PATH ?LOC-FROM ?LOC-FROM))
(DRIVE-TRUCK KEEPS (PATH ?LOC-FROM ?LOC-TO))
(DRIVE-TRUCK KEEPS (PATH ?LOC-TO ?LOC-FROM))
(DRIVE-TRUCK KEEPS (EMPTY ?TRUCK))
(DISEMBARK-TRUCK NEEDS (NOT (AT ?DRIVER ?LOC)))
(DISEMBARK-TRUCK NEEDS (DRIVING ?DRIVER ?TRUCK))
(DISEMBARK-TRUCK NEEDS (NOT (EMPTY ?TRUCK)))
(DISEMBARK-TRUCK CAUSES (AT ?DRIVER ?LOC))
(DISEMBARK-TRUCK CAUSES (NOT (DRIVING ?DRIVER ?TRUCK)))
(DISEMBARK-TRUCK CAUSES (EMPTY ?TRUCK))
(DISEMBARK-TRUCK KEEPS (AT ?TRUCK ?LOC))
(DISEMBARK-TRUCK KEEPS (PATH ?LOC ?LOC))
(BOARD-TRUCK NEEDS (AT ?DRIVER ?LOC))
(BOARD-TRUCK NEEDS (NOT (DRIVING ?DRIVER ?TRUCK)))
(BOARD-TRUCK NEEDS (EMPTY ?TRUCK))
(BOARD-TRUCK CAUSES (NOT (AT ?DRIVER ?LOC)))
(BOARD-TRUCK CAUSES (DRIVING ?DRIVER ?TRUCK))
(BOARD-TRUCK CAUSES (NOT (EMPTY ?TRUCK)))
(BOARD-TRUCK KEEPS (AT ?TRUCK ?LOC))
(BOARD-TRUCK KEEPS (PATH ?LOC ?LOC))
(UNLOAD-TRUCK NEEDS (NOT (AT ?OBJ ?LOC)))
(UNLOAD-TRUCK NEEDS (IN ?OBJ ?TRUCK))
(UNLOAD-TRUCK CAUSES (AT ?OBJ ?LOC))
```

```
(UNLOAD-TRUCK CAUSES (NOT (IN ?OBJ ?TRUCK)))
(UNLOAD-TRUCK KEEPS (AT ?TRUCK ?LOC))
(UNLOAD-TRUCK KEEPS (PATH ?LOC ?LOC))
(UNLOAD-TRUCK KEEPS (EMPTY ?TRUCK))
(LOAD-TRUCK NEEDS (AT ?OBJ ?LOC))
(LOAD-TRUCK NEEDS (NOT (IN ?OBJ ?TRUCK)))
(LOAD-TRUCK CAUSES (NOT (AT ?OBJ ?LOC)))
(LOAD-TRUCK CAUSES (IN ?OBJ ?TRUCK))
(LOAD-TRUCK KEEPS (AT ?TRUCK ?LOC))
(LOAD-TRUCK KEEPS (PATH ?LOC ?LOC))
(LOAD-TRUCK KEEPS (EMPTY ?TRUCK))
```

## C.2  Zeno-Travel Domain

The Zeno-Travel domain has the following generating PDDL:

```
(define (domain zeno-travel)
(:requirements :typing)
(:types aircraft person city flevel - object)
(:predicates (at ?x - (either person aircraft) ?c - city)
             (in ?p - person ?a - aircraft)
             (fuel-level ?a - aircraft ?l - flevel)
             (next ?l1 ?l2 - flevel))

(:action board
 :parameters (?p - person ?a - aircraft ?c - city)
 :precondition (and (at ?p ?c)  (at ?a ?c))
 :effect (and (not (at ?p ?c))  (in ?p ?a)))

(:action debark
 :parameters (?p - person ?a - aircraft ?c - city)
 :precondition (and (in ?p ?a)  (at ?a ?c))
 :effect (and (not (in ?p ?a))  (at ?p ?c)))

(:action fly
 :parameters (?a - aircraft ?c1 ?c2 - city ?l1 ?l2 - flevel)
 :precondition (and (at ?a ?c1)  (fuel-level ?a ?l1)  (next ?l2 ?l1))
 :effect (and (not (at ?a ?c1))  (at ?a ?c2)  (not (fuel-level ?a ?l1))
              (fuel-level ?a ?l2)))

(:action zoom
 :parameters (?a - aircraft ?c1 ?c2 - city ?l1 ?l2 ?l3 - flevel)
 :precondition (and (at ?a ?c1)  (fuel-level ?a ?l1)  (next ?l2 ?l1)
                 (next ?l3 ?l2) )
 :effect (and (not (at ?a ?c1))  (at ?a ?c2)  (not (fuel-level ?a ?l1))
              (fuel-level ?a ?l3)  ))

(:action refuel
 :parameters (?a - aircraft ?c - city ?l - flevel ?l1 - flevel)
 :precondition (and (fuel-level ?a ?l)  (next ?l ?l1)  (at ?a ?c))
 :effect (and (fuel-level ?a ?l1) (not (fuel-level ?a ?l)))))))
```

One learned model (one possible satisfying model of our formula) from our random-sequence input in this Zeno-Travel domain is the following (brought together with the experimental parameters):

```
Zenotravel domain:
```

```
* IPC3 problem 9
* 91 fluents, 21000 possible unique actions
* 1000 actions in learned action sequence
* 5 observed fluents per step
* "schematized" learning
* 1:1 precondition heuristics
* Action distribution: ((ZOOM . 27) (FLY . 216) (REFUEL . 264)
 (BOARD . 249) (DEBARK . 244))

converting to CNF
clause count: 71119
variable count: 138
adding clauses
calling zchaff
parsing result
SLAF time: 1.109
Inference time: 11.015
Learned model:

(REFUEL NEEDS (FUEL-LEVEL ?A ?L))
(REFUEL NEEDS (NOT (FUEL-LEVEL ?A ?L1)))
(REFUEL CAUSES (NOT (FUEL-LEVEL ?A ?L)))
(REFUEL CAUSES (FUEL-LEVEL ?A ?L1))
(REFUEL KEEPS (NEXT ?L ?L))
(REFUEL KEEPS (NEXT ?L ?L1))
(REFUEL KEEPS (NEXT ?L1 ?L))
(REFUEL KEEPS (NEXT ?L1 ?L1))
(ZOOM NEEDS (NOT (FUEL-LEVEL ?A ?L3)))
(ZOOM NEEDS (FUEL-LEVEL ?A ?L1))
(ZOOM CAUSES (FUEL-LEVEL ?A ?L3))
(ZOOM CAUSES (NOT (FUEL-LEVEL ?A ?L1)))
(ZOOM KEEPS (FUEL-LEVEL ?A ?L2))
(ZOOM KEEPS (NEXT ?L3 ?L3))
(ZOOM KEEPS (NEXT ?L3 ?L2))
(ZOOM KEEPS (NEXT ?L3 ?L1))
(ZOOM KEEPS (NEXT ?L2 ?L3))
(ZOOM KEEPS (NEXT ?L2 ?L2))
(ZOOM KEEPS (NEXT ?L2 ?L1))
(ZOOM KEEPS (NEXT ?L1 ?L3))
(ZOOM KEEPS (NEXT ?L1 ?L2))
(ZOOM KEEPS (NEXT ?L1 ?L1))
(FLY NEEDS (NOT (FUEL-LEVEL ?A ?L2)))
(FLY NEEDS (FUEL-LEVEL ?A ?L1))
(FLY CAUSES (FUEL-LEVEL ?A ?L2))
(FLY CAUSES (NOT (FUEL-LEVEL ?A ?L1)))
(FLY KEEPS (NEXT ?L2 ?L2))
(FLY KEEPS (NEXT ?L2 ?L1))
(FLY KEEPS (NEXT ?L1 ?L2))
(FLY KEEPS (NEXT ?L1 ?L1))
(DEBARK NEEDS (IN ?P ?A))
(DEBARK CAUSES (NOT (IN ?P ?A)))
(BOARD NEEDS (NOT (IN ?P ?A)))
(BOARD CAUSES (IN ?P ?A))
```

### C.3 Blocks-World Domain

The Blocksworld domain has the following generating PDDL:

```
(define (domain blocksworld)
(:requirements :strips)
(:predicates (clear ?x - object)
             (on-table ?x - object)
             (arm-empty)
             (holding ?x - object)
             (on ?x ?y - object))

(:action pickup
  :parameters (?ob - object)
  :precondition (and (clear ?ob) (on-table ?ob) (arm-empty))
  :effect (and (holding ?ob) (not (clear ?ob)) (not (on-table ?ob))
              (not (arm-empty))))

(:action putdown
  :parameters  (?ob - object)
  :precondition (holding ?ob)
  :effect (and (clear ?ob) (arm-empty) (on-table ?ob)
              (not (holding ?ob))))

(:action stack
  :parameters  (?ob - object
               ?underob - object)
  :precondition (and (clear ?underob) (holding ?ob))
  :effect (and (arm-empty) (clear ?ob) (on ?ob ?underob)
              (not (clear ?underob)) (not (holding ?ob))))

(:action unstack
  :parameters  (?ob - object
               ?underob - object)
  :precondition (and (on ?ob ?underob) (clear ?ob) (arm-empty))
  :effect (and (holding ?ob) (clear ?underob) (not (on ?ob ?underob))
              (not (clear ?ob)) (not (arm-empty)))))
```

One learned model (one possible satisfying model of our formula) from our random-sequence input in this Blocksworld domain is the following (brought together with the experimental parameters).

```
Blocksworld domain:
* 209 fluents
* 1000 randomly selected actions
* 10 fluents observed per step
* "schematized" learning
* 1:1 precondition heuristics

converting to CNF
clause count: 235492
variable count: 187
adding clauses
calling zchaff
parsing result
SLAF time: 2.203
Inference time: 42.312
```

```
Learned model:

(UNSTACK NEEDS (NOT (CLEAR ?UNDEROB)))
(UNSTACK NEEDS (CLEAR ?OB))
(UNSTACK NEEDS (ARM-EMPTY))
(UNSTACK NEEDS (NOT (HOLDING ?OB)))
(UNSTACK NEEDS (ON ?OB ?UNDEROB))
(UNSTACK CAUSES (CLEAR ?UNDEROB))
(UNSTACK CAUSES (NOT (CLEAR ?OB)))
(UNSTACK CAUSES (NOT (ARM-EMPTY)))
(UNSTACK CAUSES (HOLDING ?OB))
(UNSTACK CAUSES (NOT (ON ?OB ?UNDEROB)))
(UNSTACK KEEPS (ON-TABLE ?UNDEROB))
(UNSTACK KEEPS (ON-TABLE ?OB))
(UNSTACK KEEPS (HOLDING ?UNDEROB))
(UNSTACK KEEPS (ON ?UNDEROB ?UNDEROB))
(UNSTACK KEEPS (ON ?OB ?OB))
(UNSTACK KEEPS (ON ?UNDEROB ?OB))
(STACK NEEDS (CLEAR ?UNDEROB))
(STACK NEEDS (NOT (CLEAR ?OB)))
(STACK NEEDS (NOT (ARM-EMPTY)))
(STACK NEEDS (HOLDING ?OB))
(STACK NEEDS (NOT (ON ?OB ?UNDEROB)))
(STACK CAUSES (NOT (CLEAR ?UNDEROB)))
(STACK CAUSES (CLEAR ?OB))
(STACK CAUSES (ARM-EMPTY))
(STACK CAUSES (NOT (HOLDING ?OB)))
(STACK CAUSES (ON ?OB ?UNDEROB))
(STACK KEEPS (ON-TABLE ?UNDEROB))
(STACK KEEPS (ON-TABLE ?OB))
(STACK KEEPS (HOLDING ?UNDEROB))
(STACK KEEPS (ON ?UNDEROB ?UNDEROB))
(STACK KEEPS (ON ?OB ?OB))
(STACK KEEPS (ON ?UNDEROB ?OB))
(PUTDOWN NEEDS (NOT (CLEAR ?OB)))
(PUTDOWN NEEDS (NOT (ON-TABLE ?OB)))
(PUTDOWN NEEDS (NOT (ARM-EMPTY)))
(PUTDOWN NEEDS (HOLDING ?OB))
(PUTDOWN CAUSES (CLEAR ?OB))
(PUTDOWN CAUSES (ON-TABLE ?OB))
(PUTDOWN CAUSES (ARM-EMPTY))
(PUTDOWN CAUSES (NOT (HOLDING ?OB)))
(PUTDOWN KEEPS (ON ?OB ?OB))
(PICKUP NEEDS (CLEAR ?OB))
(PICKUP NEEDS (ON-TABLE ?OB))
(PICKUP NEEDS (ARM-EMPTY))
(PICKUP NEEDS (NOT (HOLDING ?OB)))
(PICKUP CAUSES (NOT (CLEAR ?OB)))
(PICKUP CAUSES (NOT (ON-TABLE ?OB)))
(PICKUP CAUSES (NOT (ARM-EMPTY)))
(PICKUP CAUSES (HOLDING ?OB))
(PICKUP KEEPS (ON ?OB ?OB))
```

### C.4  Depot Domain

The Depot domain has the following generating PDDL:

```
(define (domain Depot)
(:requirements :typing)
(:types place locatable - object
        depot distributor - place
        truck hoist surface - locatable
        pallet crate - surface)
(:predicates (at ?x - locatable ?y - place)
             (on ?x - crate ?y - surface)
             (in ?x - crate ?y - truck)
             (lifting ?x - hoist ?y - crate)
             (available ?x - hoist)
             (clear ?x - surface))

(:action Drive
:parameters (?x - truck ?y - place ?z - place)
:precondition (and (at ?x ?y))
:effect (and (not (at ?x ?y)) (at ?x ?z)))

(:action Lift
:parameters (?x - hoist ?y - crate ?z - surface ?p - place)
:precondition (and (at ?x ?p) (available ?x) (at ?y ?p) (on ?y ?z)
                   (clear ?y))
:effect (and (not (at ?y ?p)) (lifting ?x ?y) (not (clear ?y))
             (not (available ?x)) (clear ?z) (not (on ?y ?z))))

(:action Drop
:parameters (?x - hoist ?y - crate ?z - surface ?p - place)
:precondition (and (at ?x ?p) (at ?z ?p) (clear ?z) (lifting ?x ?y))
:effect (and (available ?x) (not (lifting ?x ?y)) (at ?y ?p)
             (not (clear ?z)) (clear ?y) (on ?y ?z)))

(:action Load
:parameters (?x - hoist ?y - crate ?z - truck ?p - place)
:precondition (and (at ?x ?p) (at ?z ?p) (lifting ?x ?y))
:effect (and (not (lifting ?x ?y)) (in ?y ?z) (available ?x)))

(:action Unload
:parameters (?x - hoist ?y - crate ?z - truck ?p - place)
:precondition (and (at ?x ?p) (at ?z ?p) (available ?x) (in ?y ?z))
:effect (and (not (in ?y ?z)) (not (available ?x)) (lifting ?x ?y)) )
```

One learned model (one possible satisfying model of our formula) from our random-sequence input in this Depot domain is the following (brought together with the experimental parameters).

```
Depots domain:
* IPC3 problem 5
* 250 fluents
* 1000 randomly selected actions
* 10 fluents observed per step
* "schematized" learning
* 1:1 precondition heuristics

converting to CNF
```

```
clause count: 85359
variable count: 236
adding clauses
calling zchaff
parsing result
SLAF time: 2.797
Inference time: 8.062
Learned model:

(UNLOAD NEEDS (IN ?Y ?Z))
(UNLOAD NEEDS (NOT (LIFTING ?X ?Y)))
(UNLOAD NEEDS (AVAILABLE ?X))
(UNLOAD CAUSES (NOT (IN ?Y ?Z)))
(UNLOAD CAUSES (LIFTING ?X ?Y))
(UNLOAD CAUSES (NOT (AVAILABLE ?X)))
(UNLOAD KEEPS (AT ?Z ?P))
(UNLOAD KEEPS (AT ?Y ?P))
(UNLOAD KEEPS (AT ?X ?P))
(UNLOAD KEEPS (ON ?Y ?Y))
(UNLOAD KEEPS (CLEAR ?Y))
(LOAD NEEDS (NOT (IN ?Y ?Z)))
(LOAD NEEDS (LIFTING ?X ?Y))
(LOAD NEEDS (NOT (AVAILABLE ?X)))
(LOAD CAUSES (IN ?Y ?Z))
(LOAD CAUSES (NOT (LIFTING ?X ?Y)))
(LOAD CAUSES (AVAILABLE ?X))
(LOAD KEEPS (AT ?Z ?P))
(LOAD KEEPS (AT ?Y ?P))
(LOAD KEEPS (AT ?X ?P))
(LOAD KEEPS (ON ?Y ?Y))
(LOAD KEEPS (CLEAR ?Y))
(DROP NEEDS (NOT (AT ?Y ?P)))
(DROP NEEDS (NOT (ON ?Y ?Z)))
(DROP NEEDS (LIFTING ?X ?Y))
(DROP NEEDS (NOT (AVAILABLE ?X)))
(DROP NEEDS (CLEAR ?Z))
(DROP NEEDS (NOT (CLEAR ?Y)))
(DROP CAUSES (AT ?Y ?P))
(DROP CAUSES (ON ?Z ?Z))
(DROP CAUSES (NOT (ON ?Z ?Z)))
(DROP CAUSES (ON ?Z ?Y))
(DROP CAUSES (NOT (ON ?Z ?Y)))
(DROP CAUSES (ON ?Y ?Z))
(DROP CAUSES (NOT (LIFTING ?X ?Y)))
(DROP CAUSES (LIFTING ?X ?Z))
(DROP CAUSES (NOT (LIFTING ?X ?Z)))
(DROP CAUSES (AVAILABLE ?X))
(DROP CAUSES (NOT (CLEAR ?Z)))
(DROP CAUSES (CLEAR ?Y))
(DROP KEEPS (AT ?Z ?P))
(DROP KEEPS (AT ?X ?P))
(DROP KEEPS (ON ?Z ?Z))
(DROP KEEPS (ON ?Z ?Y))
(DROP KEEPS (ON ?Y ?Y))
(DROP KEEPS (LIFTING ?X ?Z))
```

```
(LIFT NEEDS (AT ?Y ?P))
(LIFT NEEDS (ON ?Y ?Z))
(LIFT NEEDS (NOT (LIFTING ?X ?Y)))
(LIFT NEEDS (AVAILABLE ?X))
(LIFT NEEDS (NOT (CLEAR ?Z)))
(LIFT NEEDS (CLEAR ?Y))
(LIFT CAUSES (NOT (AT ?Y ?P)))
(LIFT CAUSES (NOT (ON ?Y ?Z)))
(LIFT CAUSES (ON ?Z ?Z))
(LIFT CAUSES (NOT (ON ?Z ?Z)))
(LIFT CAUSES (ON ?Z ?Y))
(LIFT CAUSES (NOT (ON ?Z ?Y)))
(LIFT CAUSES (LIFTING ?X ?Y))
(LIFT CAUSES (LIFTING ?X ?Z))
(LIFT CAUSES (NOT (LIFTING ?X ?Z)))
(LIFT CAUSES (NOT (AVAILABLE ?X)))
(LIFT CAUSES (CLEAR ?Z))
(LIFT CAUSES (NOT (CLEAR ?Y)))
(LIFT KEEPS (AT ?Z ?P))
(LIFT KEEPS (AT ?X ?P))
(LIFT KEEPS (ON ?Y ?Y))
(LIFT KEEPS (ON ?Z ?Z))
(LIFT KEEPS (ON ?Z ?Y))
(LIFT KEEPS (LIFTING ?X ?Z))
(DRIVE NEEDS (AT ?X ?Y))
(DRIVE NEEDS (NOT (AT ?X ?Z)))
(DRIVE CAUSES (NOT (AT ?X ?Y)))
(DRIVE CAUSES (AT ?X ?Z))
```